\documentclass[10pt,twocolumn,letterpaper]{article}

\usepackage{cvpr}              
\usepackage{graphicx}
\usepackage{amsmath}
\usepackage{amsfonts}
\usepackage{amssymb}
\usepackage{booktabs}

\usepackage[ruled,noend]{algorithm2e}
\SetKwInput{KwInput}{Inputs}                
\SetKwInput{KwOutput}{Outputs}              

\usepackage{textcomp}
\usepackage{mathtools}
\usepackage{makecell}
\usepackage{multirow}
\usepackage{epsfig}
\usepackage{amssymb}
\usepackage{mathtools}
\usepackage{multirow}
\usepackage{xcolor}
\usepackage{xspace}

\usepackage{float} 
\usepackage{stfloats}

\usepackage[pagebackref,breaklinks,colorlinks]{hyperref}

\usepackage[capitalize]{cleveref}
\crefname{section}{Sec.}{Secs.}
\Crefname{section}{Section}{Sections}
\Crefname{table}{Table}{Tables}
\crefname{table}{Tab.}{Tabs.}

\newcommand*{\rom}[1]{\expandafter\@slowromancap\romannumeral #1@}
\newcommand{\norm}[1]{\left\lVert#1\right\rVert}
\DeclareMathOperator*{\argmax}{arg\,max}

\DeclarePairedDelimiterX\set[1]\lbrace\rbrace{#1}

\newcommand{\vect}[1]{\boldsymbol{#1}}

\makeatletter
\DeclareRobustCommand\onedot{\futurelet\@let@token\@onedot}
\def\@onedot{\ifx\@let@token.\else.\null\fi\xspace}

\def\eg{\emph{e.g}\onedot} 
\def\ie{\emph{i.e}\onedot}

\def\etal{\emph{et al}\onedot}

\newcommand{\smallsym}[2]{#1{\mathpalette\make@small@sym{#2}}}
\newcommand{\make@small@sym}[2]{%
  \vcenter{\hbox{$\m@th\downgrade@style#1#2$}}%
}
\newcommand{\downgrade@style}[1]{%
  \ifx#1\displaystyle\scriptstyle\else
    \ifx#1\textstyle\scriptstyle\else
      \scriptscriptstyle
  \fi\fi
}

\newcommand{\subalign}[1]{%
  \vcenter{%
    \Let@ \restore@math@cr \default@tag
    \baselineskip\fontdimen10 \scriptfont\tw@
    \advance\baselineskip\fontdimen12 \scriptfont\tw@
    \lineskip\thr@@\fontdimen8 \scriptfont\thr@@
    \lineskiplimit\lineskip
    \ialign{\hfil$\m@th\scriptstyle##$&$\m@th\scriptstyle{}##$\hfil\crcr
      #1\crcr
    }%
  }%
}

\newcommand{\eat}[1]{}
 
\makeatother

\def\cvprPaperID{: 2830} 
\def\confName{CVPR}
\def\confYear{2022}

\begin{document}

\title{Practical Evaluation of Adversarial Robustness via Adaptive Auto Attack}

\author{Ye Liu\textsuperscript{1},
 Yaya Cheng\textsuperscript{1},
 Lianli Gao\textsuperscript{1},
 Xianglong Liu\textsuperscript{2},
 Qilong Zhang\textsuperscript{1},
 Jingkuan Song\textsuperscript{1}\thanks{Corresponding author}\\
\textsuperscript{1} Center for Future Media and School of Computer Science and Engineering\\
University of Electronic Science and Technology of China, China\\
\textsuperscript{2} Beihang University, China\\
{\tt\small \href{mailto:liuye66a@gmail.com}{liuye66a@gmail.com}, \href{mailto:yaya.cheng@hotmail.com}{yaya.cheng@hotmail.com}, \href{mailto:lianli.gao@uestc.edu.cn}{lianli.gao@uestc.edu.cn}}\\ 
{\tt\small \href{mailto:xlliu@buaa.edu.cn}{xlliu@buaa.edu.cn},
\href{mailto:qilong.zhang@std.uestc.edu.cn}{qilong.zhang@std.uestc.edu.cn},
\href{mailto:jingkuan.song@gmail.com}{jingkuan.song@gmail.com}}
}
\maketitle

\begin{abstract}
Defense models against adversarial attacks have grown significantly, but the lack of practical evaluation methods has hindered progress. 
Evaluation can be defined as looking for defense models' lower bound of robustness given a budget number of iterations and a test dataset. 
A practical evaluation method should be convenient (i.e., parameter-free), efficient (i.e., fewer iterations) and reliable (i.e., approaching the lower bound of robustness).
Towards this target, we propose a parameter-free Adaptive Auto Attack (A$^3$) evaluation method which addresses the efficiency and reliability in a test-time-training fashion.
Specifically, by observing that adversarial examples to a specific defense model follow some regularities in their starting points, we design an Adaptive Direction Initialization strategy to speed up the evaluation. 
Furthermore, to approach the lower bound of robustness under the budget number of iterations, we propose an online statistics-based discarding strategy that automatically identifies and abandons hard-to-attack images. 
Extensive experiments on nearly 50 widely-used defense models demonstrate the effectiveness of our A$^3$.
By consuming much fewer iterations than existing methods,~\ie, $1/10$ on average (10$\times$ speed up), we achieve lower robust accuracy in all cases. 
Notably, we won \textbf{first place} out of 1681 teams in CVPR 2021 White-box Adversarial Attacks on Defense Models competitions with this method. Code is available at: \href{https://github.com/liuye6666/adaptive_auto_attack}{https://github.com/liuye6666/adaptive\_auto\_attack}
\end{abstract}

\section{Introduction}
\label{sec:intro}
Despite the breakthroughs for a wide range of fields, deep neural networks (DNNs)~\cite{resnet,wider,dense,inc-v3,zhang2021curriculum} have been shown high vulnerabilities to adversarial examples. For instance, inputs added with human-imperceptible perturbations can deceive DNNs to output unreasonable predictions~\cite{cw,deepfool,color,gao2021feature,mifgsm,difgsm,nifgsm,pifgsm,pifgsm++,zhang2022bia,hit}. 
To tackle this issue, various adversarial defense methods~\cite{mask1,defense_auxilary2,defense_preprocess2,adversarial_train1} have been proposed to resist against malicious perturbations. Unfortunately, these defense methods could be broken by more advanced attack methods~\cite{failed_defense2,failed_defense3,failed_defense4,aa}, making it difficult to identify the state-of-the-art. Therefore, we urgently need a practical evaluation method to judge the adversarial robustness of different defense strategies. 

Robustness evaluation can be defined as looking for defense models' lower bound of robustness given a budget number of iterations and a test dataset~\cite{aa}. 
White-box adversarial attacks on defense models are crucial for testing adversarial robustness. 
Among these methods, widely-used random sampling has been proven effective in generating diverse \textbf{starting points} for attacks in a large-scale study~\cite{pgd,odi,aa,mt-pgd}. 
In general, there are two kinds of random sampling strategies. 
From the perspective of input space, given an original image $\vect x$ of label $y$ and a random perturbation $\vect \zeta$ sampled from uniform distributions, the starting point is $\vect{x_{st}} = \vect{x} + \vect \zeta$,~\eg, Projected Gradient Descent (PGD)~\cite{pgd}. 
From the perspective of output space, given a classifier $f$ and a randomly sampled direction of diversification $\vect{w_d}$, evaluators generate starting points by maximizing the change of output,~\ie, $\vect{w_d^\intercal}f(\vect{x})$. 
Intuitively, random sampling strategy is sub-optimal since it is model-agnostic. 

Comprehensive statistics on random sampling are conducted to verify whether it is sub-optimal. In other words, we want to study whether adversarial examples,~\ie, images that successfully fooling the victim's models, to a specific defense model follow some regularities in their starting points.
Statistic results on input space show that adversarial examples' starting points are actually random. This is reasonable because starting points are highly dependent on their corresponding input images, and input images are randomly distributed in high-dimensional space. 
Different from input space, statistics results in output space show that the direction of diversification $\vect{w_d}$ to a specific defense model follows some regularities.
Specifically, $\vect{w_d}$ is not uniformly distributed but with a model-specific bias in the positive/negative direction.
Therefore, random sampling in the output space cannot obtain a good starting point, which may slow down the evaluation. To speed up the evaluation, we propose an Adaptive Direction Initialization (ADI) strategy in this paper. ADI firstly adopts an observer to record the direction of diversification of adversarial examples at the first restart. 
Then, based on these directions, ADI introduces a novel way to generate better staring points than random sampling for the following restarts. 

In addition to using ADI to accelerate robustness evaluation, we design another strategy named online statistics-based discarding for improving the reliability of existing methods. 
Currently, the na{\"i}ve iterative strategy that treats all images evenly and allocates them the same iterations is widely applied to robustness evaluation~\cite{aa,pgd,odi,fab,mt-pgd,CAA,lafeat}. 
However, this strategy is unreasonable because it pays unnecessary efforts to perturb hard-to-attack images.
Intuitively, given the budget number of iterations, the more examples we successfully attack, the closer robustness to the lower bound we obtain. 
Therefore, the number of iterations assigned to hard-to-attack images is a lower priority. 
Based on our observation that loss values can roughly distinguish the difficulty of attacks, we propose an online statistics-based discarding strategy that automatically identifies and abandons hard-to-attack images. Specifically, we stop perturbing images with considerable difficulties at the beginning of every restart. For remaining images, the same number of iterations are allocated to them. Obviously, online statistics-based discarding strategy makes full use of the number of iterations and increases the chance of perturbing images to adversarial examples.
We can further approach the lower bound of robustness based on this reliable strategy. 
Essentially, speeding up the evaluation is also closely related to improving the reliability because the saved iterations can be used to attack easy-to-attack examples, resulting in  lower robust accuracy.
By incorporating the above two strategies, a practical evaluation method Adaptive Auto Attack (A$^3$) is proposed.

To sum up, our main contributions are three-fold: 
1) Based on comprehensive statistics, we propose an adaptive direction initialization (ADI) strategy which generates better starting points than random sampling to speed up the robustness evaluation. 2) We propose an online statistics-based discarding strategy that automatically identifies and abandons hard-to-attack images to approach further the lower bound of robustness  under the budget number of iterations. 3) Extensive experiments demonstrate the effectiveness and reliability of the method. Particularly, we apply A$^3$ to nearly 50 widely-used defense models, without parameters adjustment, our method achieves a lower robust accuracy and a faster evaluation. 

\section{Related Works}

Many white-box attacks against defense models have been proposed for robustness evaluation. 
Fast Adaptive Boundary Attack (FAB)~\cite{fab} aims at finding the minimal perturbation necessary to change the class of a given input. 
Projected Gradient Descent (PGD)~\cite{pgd} further improves the evaluation performance by assigning random perturbations as the starting point at every restart. 
Based on PGD, Gowal~\etal~\cite{mt-pgd} proposes a MultiTargeted attack (MT) that picks a new target class at each restart.
Tashiro~\cite{odi} provides a more effective initialization strategy to generate diverse starting points. 
Unfortunately, 
most of these promising methods overestimate the robustness~\cite{failed_defense1,failed_defense2,failed_defense3,failed_defense4}. 
Potential reasons for this are improper hyper-parameters tuning and gradient masking~\cite{aa}. 
Therefore, we urgently need a practical evaluation method that is convenient (~\ie, parameter-free), efficient (~\ie, less iterations), and reliable (~\ie, approaching the lower bound of robustness).

To address this issue, Croce~\etal~\cite{aa} proposes Auto Attack (AA) by integrating four attacks methods. 
Large-scale studies have shown that AA achieves lower robust test accuracy than existing methods. 
However, AA is inefficient since it requires a massive number of iterations for robustness evaluation. 
Instead of adopting an ensemble strategy, Yu~\etal~\cite{lafeat} uses latent features and introduces a unified $\ell_\infty$-norm white-box attack algorithm LAFEAT to evaluate robustness more reliably. 
However, the time and space complexity of LAFEAT is unacceptable, as it requires the training of new modules for each defense model. 
Generally speaking, it is impracticable since training sets are often inaccessible in practical applications.

\section{Methodology}
\subsection{Preliminaries}
\label{sec:formatting}
In this section, we give the background knowledge of adversarial attacks.
Given a $C$-class classifier 
$f:\vect{x} \in [0, 1]^D\rightarrow \mathbb{R}^C$, where $\vect{x}$ is original image with label $y$, model prediction is calculated by: 
\begin{equation}
 h(\vect {x}) = \underset{c=1,...,C}{\mathrm{argmax}}\, f_c(\vect {x}),
\end{equation}
where $f_c(\vect {x})$ refers to the output logits of $\vect x$ on the $c$-th class. In this paper, we mainly focus on untargeted attacks. The goal of untargeted adversarial attacks is fooling $f$ to missclassify a human-imperceptible adversarial example $\vect{x_{adv}} = \vect x + \vect \delta$,~\ie, $ h(\vect {x_{adv}}) \neq y$. By adopting $\ell_\infty$ distance to evaluate the imperceptibility of $\vect{\delta}$,~\ie, $\norm{\vect \delta}_{\infty}\leq \epsilon$, the constrained optimization problem is defined as:
\begin{align}
{\argmax}\,\mathcal{L}\left(f\left(\vect{x_{adv}}\right) ,y \right) \quad \mathit{s.t.} \norm{\vect{x_{adv}} - \vect{x}}_\infty \leq \epsilon.
\end{align}

In the scenario of white-box adversarial attacks, attackers can access all information of the victims' model. 
In this case, one of the most popular methods is PGD~\cite{pgd}. Specifically, PGD calculates the gradient at iteration $t$:
\begin{align}
\vect{g^t}= \nabla_{\vect {x_{adv}^t}} \mathcal{L}\left(f\left(\vect {x_{adv}^t}\right) ,y \right),
\end{align}
where $\vect {x_{adv}^t}$ is the adversarial example at iteration $t$ and the starting point $\vect {x_{st}}$ is generated as:
\begin{align}
\vect {x_{st}} =\vect{x}+\vect \zeta,
\label{eq:pgd_startpoint}
\end{align}
where $\vect \zeta$ is a random perturbation sampled from a uniform distribution $U\left(-\epsilon, \epsilon \right)^{D}$. Then, PGD generates an adversarial example by performing the iterative update:
\begin{align}
\label{eq_pgd}
\vect{x_{adv}^{t+1}}=P_{\vect {x}, \epsilon}\left(\vect{x_{adv}^t}+\eta ^t \cdot sign( \vect {g^t}) \right),
\end{align}
where $\eta ^t$ is the step size at iteration $t$ and $\vect{x}_{\vect{adv}}^{0}=\vect{x_{st}}$, function $P_{\vect {x}, \epsilon}\left(\cdot\right)$ clips the input to the $\epsilon$-ball of $\vect {x}$. To accurately evaluate the robustness of defense models, PGD usually adopts multiple restarts. At each restart, starting points are randomly sampled from the perturbation space.
 
To further improve the diversity of starting points, Tashiro \etal~\cite{odi} propose ODI to find starting points by maximizing the change in the output space. To be specific, given a random direction of diversification $\vect{w_{d}}$ sampled from uniform distributions $U\left(-1, 1 \right)^{C}$, ODI firstly calculates a normalized perturbation vector as follows:
\begin{align}
    \vect{\upsilon}\left(\vect x, f, \vect{w_d} \right) = \frac{\nabla_{\vect x}{\vect{w_d^\intercal}f\left(\vect x\right)}}{\norm{\nabla_{\vect x}{\vect{w_d^\intercal}f\left(\vect x\right)}}},
\end{align}
and then generates starting points by maximizing the output change via the following iterative update:
\begin{align}
    \begin{split}
        \vect {x^{t+1}_{adv}} &= \vect {x^t_{adv}} + \eta _{odi} \cdot \operatorname{sign}\left(\vect{\upsilon}\left(\vect{x}_{\vect{adv}}^t, f, \vect{w_d} \right)\right),\\
        \vect{x_{adv}^{t+1}} &= P_{\vect x,\epsilon}\left(\vect{x_{adv}^{t+1}}\right),
    \end{split}
    \label{eq_odi}
\end{align}
where $\eta _{odi}$ is the step size for ODI, which is usually set to $\epsilon$, and in this paper we keep the same setting.
$\vect{x}_{\vect{adv}}^{0}$ is calculated by~\cref{eq:pgd_startpoint}. After $N_{odi}$ iterations,~\ie, number of iteration for initialization, ODI obtains the starting point $\vect{x_{st}}$:
\begin{align}
\vect{x_{st}} = \vect{x_{adv}^{N_{odi}}}.
\label{eq:odi_startpoint}
\end{align}
As same as PGD, ODI performs multiple restarts to achieve lower robust accuracy.

\subsection{Motivations}
\label{sec:motivation}
A practical evaluation method should be convenient (~\ie, parameter-free\footnote{Following  \cite{aa}, `parameter-free' indicates that we do not need the fine-tuning of parameters for every new defense.}), efficient (~\ie, fewer iterations), and reliable (~\ie, approaching the lower bound of robustness). Although using a large number of iterations, most of the existing methods usually overestimate the robustness. There are two potential reasons for this: a) Despite the effectiveness of generating diverse starting points for attacks, random sampling is sub-optimal since it is model-agnostic. Exploiting random sampling to generate starting points will slow down the robustness evaluation, and b) The widely adopted na{\"i}ve iterative strategy,~\ie, assigning the same number of iteration to all test examples, is unreasonable. Intuitively, na{\"i}ve iterative  strategy pays unnecessary efforts to perturb hard-to-attack images.

To verify the above two points, we perform a comprehensive statistical analysis of some white-box adversarial attacks methods against defense models.
\begin{figure}[t]
\centering
\includegraphics[width=0.4\textwidth]{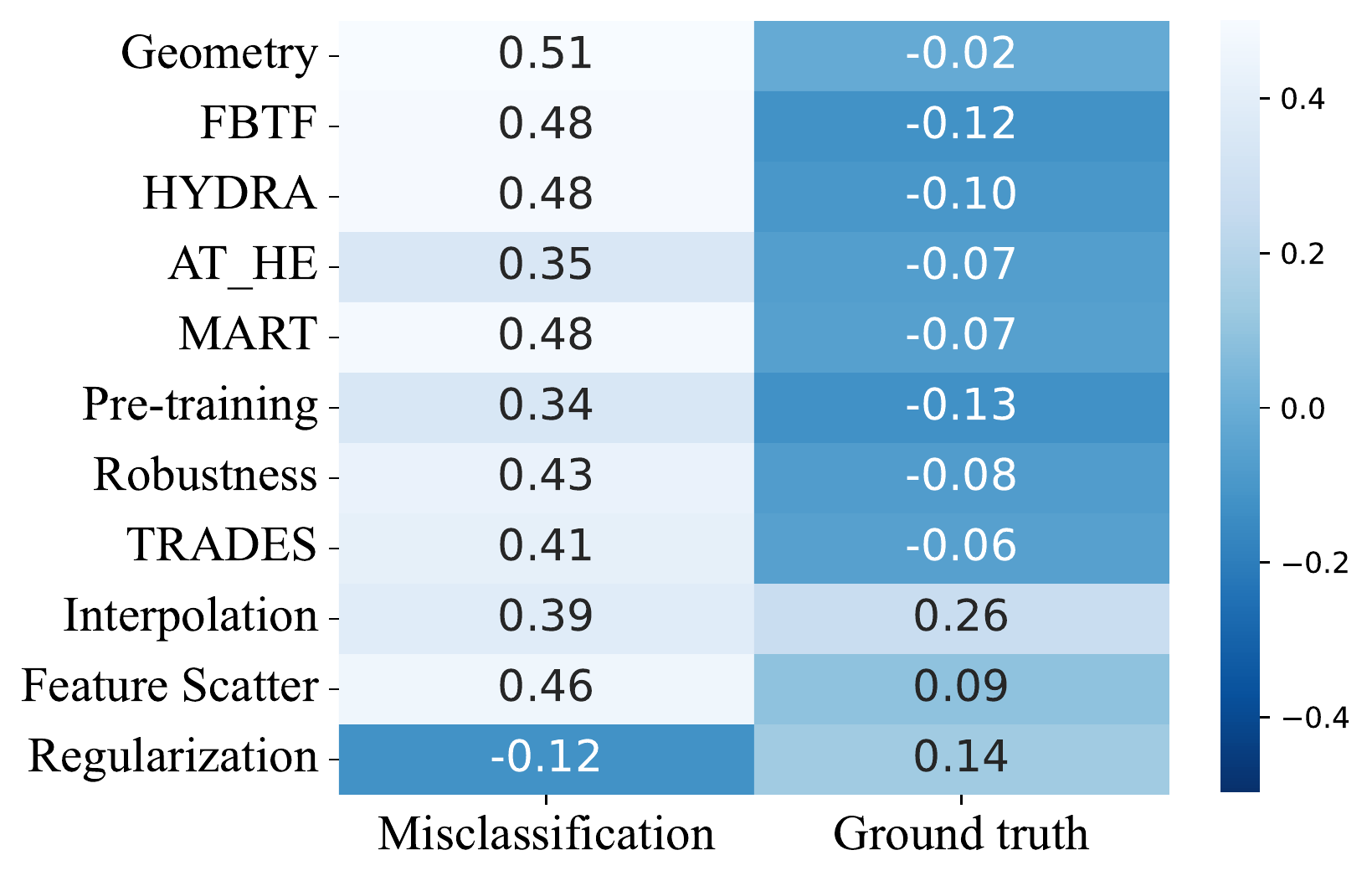}
\vspace{-0.7em}
\caption{Quantitative statistical results of the diversified direction $\vect w_{\vect d}$ of adversarial examples on 11 models.
The diversified direction $\vect w_{\vect d}$ for all models disobey uniform distribution. 
} 
\vspace{-0.7em}
\label{fig:sta_odi}
\end{figure}

\noindent{\textbf{Random sampling is sub-optimal.}} Considering \cref{eq:odi_startpoint} and \cref{eq:pgd_startpoint}, random sampling is widely used to generate diverse starting points for  attacks in the input space (~\eg, PGD), and output space (~\eg, ODI). It is worth noting that there is no need to do statistics on random sampling in the input space, since the starting points are highly dependent on the corresponding input images and the input images are randomly distributed in the high-dimensional space of the input space. 
Therefore, we mainly focus on the output space. 

As shown in \cref{eq_odi}, the noise in the output space is determined by $\vect{w_d}$. Therefore, to verify the unreasonableness of random sampling in the output space, we analyze $\vect{w_d}$ of adversarial examples (\ie, examples that were successfully attacked) and explore what kind of $\vect{w_d}$ is conducive. 
Specifically, 
we adopt ODI parameterized by number of restarts $R=50$, $N_{odi}=7$ and $\eta _{odi}=\epsilon$. The number of iterations for attacks at each restarts $N_{atk}$ is set to 30, and the step size at $t$-th iteration is formulated as follows:
\begin{equation}
    \eta^t =  \frac{1}{2}\epsilon \cdot \left( 1+cos\left( \frac{t\ {\rm mod} \ N_{atk}}{N_{atk}} \pi \right) \right).
    \label{eq: step size} 
\end{equation}
Then we use ODI to attack 11 defense models, including Geometry~\cite{Geometry}, FBTF~\cite{fbtf}, HYDRA~\cite{HYDRA}, AT\_HE~\cite{Hypersphere}, MART~\cite{MARTs}, Pre-training~\cite{Pre-training}, Robustness~\cite{robustness}, TRADES~\cite{TRADESs}, Interpolation~\cite{Interpolation}, Feature Scatter~\cite{Feature-scatter}, Regularization~\cite{Regualr}. 
Among adversarial examples against different models, we summarize statistic results of $\vect{w_d}$ in \cref{fig:sta_odi}. The 1st and 2nd columns give mean values of $\vect{w_d}$ at the $\hat{y}$-th (the misclassification label),~\ie, $\overline{\vect{w}_{\vect{d}}^{\hat y}}$, and the $y$-th (the ground truth),~\ie, $\overline{\vect w_{\vect d}^y}$. 

From \cref{fig:sta_odi}, we have the following observations. Firstly, for adversarial examples, their direction of diversification $\vect{w_d}$ disobeys uniform distribution. 
Secondly, there is a model-specific positive/negative bias in the direction of diversification. 
Mainly, there are three kinds of biases: \textbf{a}) $\overline{\vect w_{\vect d}^y}<0$, $\overline{\vect{w}_{\vect{d}}^{\hat y}}>0$, \textbf{b}) $\overline{\vect w_{\vect d}^y}>0$, $\overline{\vect{w}_{\vect{d}}^{\hat y}}>0$, and \textbf{c}) $\overline{\vect w_{\vect d}^y}>0$, $\overline{\vect{w}_{\vect{d}}^{\hat y}}<0$. Considering \cref{eq_odi}, for \textbf{a}, $f_y\left({\vect {x_{st}}}\right)$ and $f_{\hat{y}}\left({\vect {x_{st}}}\right)$ will decrease and increase respectively, which satisfies the goal of adversarial attacks. For \textbf{b}, both of $f_y\left({\vect {x_{st}}}\right)$ and $f_{\hat{y}}\left({\vect {x_{st}}}\right)$ will increase. For \textbf{c}, $f_y\left({\vect {x_{st}}}\right)$ will increase and $f_{\hat{y}}\left({\vect {x_{st}}}\right)$ will decrease. 
Obviously, cases \textbf{b} and \textbf{c} are counter-intuitive, a potential reason is that these defense models adopt gradient masks~\cite{mask}. Based on these observations, random sampling is sub-optimal since it is model-agnostic. Adopting it to generate starting points hinders the algorithms from approaching the lower bound of robustness rapidly. 
For more detailed statistical results of $\vect{w_d}$, please refer to the Appendix. B.

\noindent{\textbf{Limitations of Na{\"i}ve iterative  strategy.}} Most of the existing methods adopt the na{\"i}ve iterative strategy,~\ie, treats all images evenly. However, based on two intuitions: (1) Images vary in difficulty to perturb them to adversarial examples, and (2) The higher the difficulty, the more iterations are needed to perturb them. Na{\"i}ve iterative  strategy is impractical because it pays unnecessary efforts to perturb hard-to-attack images. In order to successfully attack more images and get closer to the lower bound of robustness with the budget number of iterations, the number of iterations assigned to hard-to-attack images is a lower priority.
Therefore, we need a method that can roughly distinguish hard-to-attack images and easy-to-attack images to allocate the budget number of iterations reasonably.

Intuitively, loss function values can roughly reflect the difficulty of perturbing an image to an adversarial example. 
Multiple loss functions $\mathcal{L(\cdot)}$ can be used for attacks, including cross-entropy loss and margin loss defined as $max_{c\neq  y} f_{c} (\vect x)-f_{y} (\vect x)$. In this paper, we use the margin loss and define hard-to-attack images as images that cannot be successfully attacked even after 2000 iterations. The other images that are successfully attacked, we define as easy-to-attack images.

\begin{figure}[t]
\centering
\includegraphics[width=0.45\textwidth]{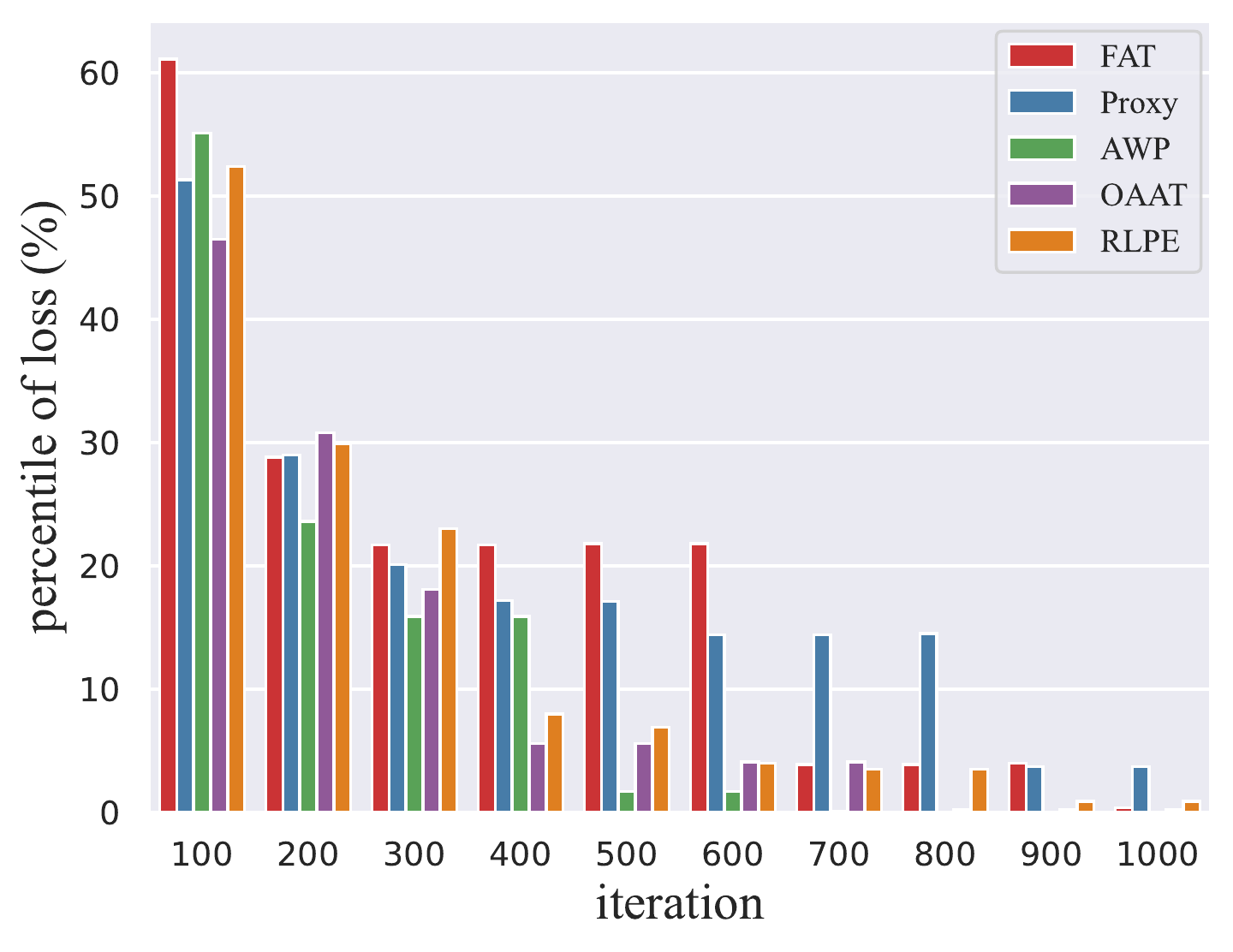} 
\vspace{-0.7em}
\caption{Quantitative statistical results of loss percentile of easy-to-attack images. As the number of iterations increases, the loss percentile of easy-to-attack images continually decrease.}
\vspace{-0.7em}
\label{fig:iteration}
\end{figure}

To verify that loss values can distinguish between hard-to-attack and easy-to-attack images, given 2000 iterations for attacks, we first use ODI to attack all images of 5 models (including, AWP~\cite{AWP}, FAT~\cite{FAT}, Proxy~\cite{Proxy}, OAAT~\cite{OAAT} and RLPE~\cite{RLPE}), and then identify and mark the easy-to-attack images of each model. Finally, we attack the 5 models again with ODI to record the percentile of loss values in descending order for easy-to-attack images in the process of attacks. 
We visualized the statistical results in \cref{fig:iteration}. 

From this figure, we have the following observations. 
As the number of iterations increases, the loss percentiles of easy-to-attack images continually decrease. Besides, the loss percentiles of easy-to-attack images always take a higher position than that of most hard-to-attack images. Take easy-to-attack image as an example, when the number of iterations reaches 100, the percentile of loss ranks in the top 60\%, as the number of iterations increases, it decreases to the top 5\% or even the top 0.1\%.
In other words, loss values of easy-to-attack and hard-to-attack images are not randomly distributed, and the loss values of hard-to-attack images are more likely to be small. We can distinguish between these images according to loss value. Based on these observations, to make full use of the budget number of iterations, we can automatically abandon hard-to-attack images with an increasing proportion according to loss values during the process of attacks.

\subsection{Adaptive Direction Initialization}
Inspired by the above analysis of random sampling in Sec.~\ref{sec:motivation}, we propose a method named Adaptive Direction Initialization (ADI) to generate better directions than random sampling to initialize the attacks. Specifically, ADI has two steps: 
\textit{useful directions observer} and \textit{adaptive directions generation}. 

For \textit{useful directions observer} step, ADI first adopts random sampling to generate the direction of diversification,~\ie, $\vect{w_d}=U\left(-1,1\right)^{C}$. Then ADI uses starting points obtained by \cref{eq:odi_startpoint} to initialize PGD attacks and obtains adversarial examples crafted by PGD. 
We denote $W$ as a set containing $\vect{w_d}$ of all adversarial examples.

Motivated by Sec.~\ref{sec:motivation}, for \textit{adaptive directions generation}, ADI adopts the sign of the summed $\vect{w_d}$ in $W$ as prior knowledge to generate the adaptive direction $\vect{w_a}$:
\begin{equation}
\label{d_sign}
\vect{\kappa}_{c}(W) = sign\left(\begin{matrix} \sum\limits_{\vect{w_{d}} \in W}{\vect{w}_{\vect{d}}^c}\end{matrix}\right),
\end{equation}
where $\vect{\kappa}_{c}(W)$ is the prior knowledge of generating $\vect{w}_{\vect a}^c$,~\ie, the $c$-th dimension of $\vect{w_a}$. With the help of $\vect{\kappa}(\cdot)$, ADI generates the $y$-th components of $\vect{w_a}$ as:
\begin{align}
\label{eq:w1}
\begin{split}
    \vect{w}_{\vect a}^{y}=&\left\{\begin{array}{ll}
    \vect{w}_{\vect a}^{y} \sim U\left( -0.5,0.1 \right), &\vect{\kappa}_{y}(\cdot)< 0,\\
    \vect{w}_{\vect a}^{y} \sim U\left(-0.1,0.5\right),  &\vect{\kappa}_{y}(\cdot)> 0.
    \end{array}\right.
\end{split}
\end{align}
To improve the effectiveness of $\vect{w}_{\vect a}$, ADI randomly selects a label $\mathfrak y$ to follow the sign of symbol of $\vect{\kappa}_{\mathfrak{y}}(\cdot)$:
\begin{align}
\label{eq:w2}
    \vect{w}_{\vect a}^{\mathfrak{y}}=&\left\{\begin{array}{ll}
    -0.8,    &\vect{\kappa}_{\mathfrak{y}}(\cdot)< 0, \\
    0.8,     &\vect{\kappa}_{\mathfrak{y}}(\cdot)> 0.
    \end{array}\right.
\end{align}
Notably, our method is insensitive to $\vect{w}_{\vect a}^{\mathfrak{y}}$ experimentally. For simplicity, we set $\vect{w}_{\vect a}^{\mathfrak{y}}=\pm0.8$. For the remaining dimensions of adaptive direction $\vect{w}_{\vect a}$, ADI calculates them as: 
\begin{align}
\label{eq:w3}
     \vect{w}_{\vect a}^{i} \sim U\left(-1,1\right)^{C-2}, ~ i\in \mathcal{T},
\end{align}
where the set $\mathcal{T}=\{1,...,C\}\backslash \{y,\mathfrak y\}$ contains all classes other than $y$ and $\mathfrak y$.
Compared with random sampling adopted by ODI, the adaptive direction generated by ADI is guided by prior knowledge,~\ie, the direction of diversification of adversarial examples. Furthermore, ADI randomly generates the remaining $C-2$ dimensions of the adaptive direction to improve the diversity of starting points. 
\begin{algorithm}[!t]
    \caption{Adaptive Auto Attack (A$^{3}$)}
    \label{alg:AAA}
    \SetNoFillComment
    \SetInd{0.2em}{0.6em}
    \KwInput{norm bound $\epsilon$, the number of iteration for initialization $N$, step sizes $\eta$, the number of iteration for attacks at the $r$-th restart $N^{r}_{atk}$, attack iteration step sizes $\eta _{atk}$, number of restarts $R$, test dataset $\mathcal{I}$}
    \KwOutput{Adversarial example $\vect{x}_{\vect{adv}}^{t+1}$}

    
    \For{$r=0 \rightarrow R$}{
        Update the test dataset $\mathcal{I}$ by OSD
        
        \For{$\vect x$ in $\mathcal{I}$}{
            Sample $\vect \zeta$ from $U\left(-\epsilon, \epsilon \right)^{D}$
            
            $\vect{x_{st}} = \vect x+ \zeta$
            
            \If{{$r=0$}}{
                Sample $\vect w_{\vect d}$ from $U\left(-1, 1 \right)^{C}$
            }
            \Else{
                $\vect w_{\vect d} \gets \vect w_{\vect a}$
            }
            \tcc{ADI}
            \For{$n=0$ to $N$}{
                Compute $\vect{x}_{\vect{adv}}^{n+1}$ by \cref{eq_odi}
            }
            \For{$t=0$ to $N^{r}_{atk}$}{
               Compute $\vect{x}_{\vect{adv}}^{t+1}$ by \cref{eq_pgd} 
            }
            \If{r=0}{
                Compute $\vect w_{\vect a}$ by \cref{eq:w1,eq:w2,eq:w3}.
            }
            \If{$\vect{x}_{\vect{adv}}^{t+1}$ is an adversarial example}{
                \Return{ $\vect{x}_{\vect{adv}}^{t+1}$}
            }

        }
    }

\end{algorithm}
\subsection{Online Statistics-based Discarding Strategy}
In light of approaching the lower bound of robustness with the budget number of iterations, we propose a novel iterative strategy named Online Statistics-based Discarding (OSD). 
According to the observation in Sec.~\ref{sec:motivation}. OSD adopts loss values to distinguish between hard-to-attack and easy-to-attack images. 
OSD first sorts test images in descending order by the corresponding loss values at the beginning of every restart, 
and then discards hard-to-attack images, ~\ie, stopping perturbing images with small loss values.
Particularly, given an initial discarding rate $\phi$ and an discarding increment $\iota$, the discarding rate at the $r$-th restart is formulated as follows:
\begin{align}
\label{eq: discarding rate}
    \varsigma^r = \phi + r\times\iota.
\end{align}
For the remaining images, OSD assigns the same number of iterations to them. Intuitively, to further increase attack success rate, OSD allocates more iterations to remaining images at restart $r$ than previous restart. Concretely, given an initial number of iteration $\gamma$ for attacks and an iteration increment $\nu$, the number of iterations for attacks at the $r$-th restart is computed as follows:
\begin{align}
    N_{atk}^r = \gamma + r\times\nu.
\end{align}

Compared with na{\"i}ve iterative  strategy, OSD makes full use of the budget number of iterations by automatically identifying and abandoning hard-to-attack images. 
In addition, by allocating various number of iterations for attacks at different restarts, OSD helps to further approach the lower bound of adversarial robustness. 

\subsection{Adaptive Auto Attack}
We integrate above two strategies to form a practical evaluation method Adaptive Auto Attack (A$^3$).
Firstly, A$^3$ is convenient since we do not need the fine-tuning of parameters for every new defense models.
Secondly, A$^3$ is efficient. 
For the gradient-based optimization to craft adversarial examples, unlike random sampling, 
our method A$^3$ generates adaptive directions for each model and provides better starting points to speed up the evaluation. Thirdly, A$^3$ is reliable. By discarding hard-to-attack images online and adjusting iterations for attacks adaptively, our method A$^3$ makes full use of a budget number of iterations and further approaches the lower bound of adversarial robustness. Compared with the mainstream method AA, our parameter-free A$^3$ is a more efficient and reliable protocol for robustness evaluation. 
The algorithm of Adaptive Auto Attack is summarized in Algorithm~\ref{alg:AAA}. 

\section{Experiments}
We conduct comprehensive experiments to evaluate the practicability of our method. Specifically, five baselines are included: PGD~\cite{pgd}, ODI~\cite{odi}, MT-PGD~\cite{mt-pgd}, I-FGSM~\cite{ifgsm} and AA~\cite{aa}. 
Nearly 50 $\ell_\infty$-defense models with 8 different architectures are chosen from recent conferences. 
To be specific, the evaluation is performed on 35 and 12 defense models trained on CIFAR-10 and CIFAR-100~\cite{cifar-10} datasets, respectively. 
Notably, for fair comparisons, the step size is calculated by~\cref{eq: step size} for all attack methods in the experiments. We use margin loss for all attack methods. 

Following AA, we adopt robust accuracy~(\textbf{acc}) to reflect evaluation reliability. 
A robustness evaluation method is considered reliable if it can better downgrade the model's classification accuracy.
In our experiments, we assume computational complexity of all methods in each iteration is similar. So the evaluation efficiency of each method can be reflected by the total number of iterations. For simplicity, we test the forward propagation~(``\textbf{$\rightarrow$}") and backward propagation~(``\textbf{$\leftarrow$}") to indicate the evaluation efficiency of different methods. 
\begin{table*}[t]
  \centering
  \resizebox{\textwidth}{!}{
  \begin{tabular}{@{}r|cccccccccccc@{}}\toprule
  \multirow{2}{*}{\textbf{\makecell[c]{CIFAR-10 \\Defense Method}}}&\multirow{2}{*}{\textbf{Model}}&\textbf{Clean}&\textbf{Nominal}&\textbf{PGD}&\textbf{ODI}&\multicolumn{3}{c}{\textbf{AA}}&\multicolumn{3}{c}{\textbf{A$^{3}$}}&$\bf{\Delta}$\\\cmidrule(lr){3-3}\cmidrule(lr){4-4}\cmidrule(lr){5-5}\cmidrule(lr){6-6}\cmidrule(lr){7-9}\cmidrule(lr){10-12}\cmidrule(l){13-13}
                                   &    & acc  &acc &acc   & acc  & acc &$\rightarrow$&$\leftarrow$& acc &$\rightarrow$&$\leftarrow$& acc    \\\midrule
  ULAT\cite{ULAT}$^\dagger$     &WRN-70-16&$91.10$&$65.87$&$66.75$&$66.06$&$65.88$&$51.20$&$12.90$&$\bf{65.78\downarrow0.10}$&$\bf{4.49(11.40\times)}$&$\bf{2.20(5.86\times)}$&$\downarrow0.09$\\
  Fixing Data\cite{fix_data}      &WRN-70-16&$88.54$&$64.20$&$65.10$&$64.46$&$64.25$&$50.82$&$12.59$&$\bf{64.19\downarrow0.06}$&$\bf{4.41(11.52\times)}$&$\bf{2.17(5.81\times)}$&$\downarrow0.01$\\
  ULAT\cite{ULAT}$^\dagger$     &WRN-28-10&$89.48$&$62.76$&$63.63$&$63.01$&$62.80$&$49.62$&$12.30$&$\bf{62.70\downarrow0.10}$&$\bf{4.28(11.58\times)}$&$\bf{2.10(5.85\times)}$&$\downarrow0.05$\\
  Fixing Data\cite{fix_data}      &WRN-28-10&$87.33$&$60.73$&$61.64$&$61.09$&$60.75$&$47.98$&$11.91$&$\bf{60.66\downarrow0.09}$&$\bf{4.14(11.59\times)}$&$\bf{2.04(5.83\times)}$&$\downarrow0.07$\\
  RLPE\cite{RLPE}$^\dagger$               &WRN-34-15&$86.53$&$60.41$&$61.25$&$60.69$&$60.41$&$47.53$&$11.82$&$\bf{60.31\downarrow0.10}$&$\bf{4.12(11.52\times)}$&$\bf{2.02(5.84\times)}$&$\downarrow0.10$\\
  AWP\cite{AWP}$^\dagger$       &WRN-28-10&$88.25$&$60.04$&$60.55$&$60.23$&$60.04$&$47.20$&$11.70$&$\bf{59.98\downarrow0.06}$&$\bf{4.09(11.54\times)}$&$\bf{2.01(5.82\times)}$&$\downarrow0.06$\\
  RLPE\cite{RLPE}$^\dagger$     &WRN-28-10&$89.46$&$59.66$&$60.78$&$59.88$&$59.66$&$47.09$&$11.72$&$\bf{59.51\downarrow0.15}$&$\bf{4.10(11.49\times)}$&$\bf{2.00(5.85\times)}$&$\downarrow0.15$\\
  Geometry\cite{Geometry}$^{\dagger\ddagger}$&WRN-28-10&$89.36$&$59.64$&$60.17$&$59.59$&$59.64$&$47.10$&$11.67$&$\bf{59.53\downarrow0.11}$&$\bf{4.10(11.49\times)}$&$\bf{2.00(5.85\times)}$&$\downarrow0.11$\\
  RST\cite{RST}$^\dagger$       &WRN-28-10&$89.69$&$62.50$&$60.64$&$59.44$&$59.53$&$47.10$&$11.70$&$\bf{59.42\downarrow0.11}$&$\bf{4.10(11.49\times)}$&$\bf{2.01(5.82\times)}$&$\downarrow3.08$\\
  Proxy\cite{Proxy}$^\dagger$   &WRN-34-10&$85.85$&$59.09$&$60.51$&$59.94$&$59.09$&$46.70$&$11.60$&$\bf{58.99\downarrow0.10}$&$\bf{4.04(11.56\times)}$&$\bf{1.98(5.86\times)}$&$\downarrow0.10$\\
  OAAT\cite{OAAT}               &WRN-34-10&$85.32$&$58.04$&$58.84$&$58.25$&$58.04$&$45.64$&$11.34$&$\bf{57.98\downarrow0.06}$&$\bf{3.99(11.43\times)}$&$\bf{1.96(5.76\times)}$&$\downarrow0.06$\\
  HYDRA\cite{HYDRA}$^\dagger$   &WRN-28-10&$88.98$&$59.98$&$58.27$&$57.60$&$57.14$&$45.20$&$11.20$&$\bf{57.06\downarrow0.08}$&$\bf{3.91(11.56\times)}$&$\bf{1.92(5.83\times)}$&$\downarrow2.92$\\
  ULAT\cite{ULAT}               &WRN-70-16&$85.29$&$57.20$&$57.90$&$57.48$&$57.20$&$45.20$&$11.20$&$\bf{57.08\downarrow0.12}$&$\bf{3.90(11.59\times)}$&$\bf{1.92(5.83\times)}$&$\downarrow0.12$\\
  ULAT\cite{ULAT}               &WRN-34-20&$85.64$&$56.82$&$57.40$&$57.00$&$56.86$&$44.96$&$11.18$&$\bf{56.76\downarrow0.10}$&$\bf{3.88(11.60\times)}$&$\bf{1.90(5.89\times)}$&$\downarrow0.10$\\
  MART\cite{MARTs} $^\dagger$   &WRN-28-10&$87.50$&$65.04$&$58.09$&$56.80$&$56.29$&$44.60$&$11.10$&$\bf{56.20\downarrow0.09}$&$\bf{3.86(11.55\times)}$&$\bf{1.89(5.93\times)}$&$\downarrow8.84$\\
  Pre-training\cite{Pre-training}$^\dagger$ &WRN-34-10&$87.11$&$57.40$&$56.43$&$55.32$&$54.92$&$43.40$&$10.80$&$\bf{54.76\downarrow0.16}$&$\bf{3.73(11.64\times)}$&$\bf{1.83(5.90\times)}$&$\downarrow2.64$\\
  Proxy\cite{Proxy}             &ResNet-18&$84.38$&$55.60$&$56.31$&$54.98$&$54.43$&$43.21$&$10.71$&$\bf{54.35\downarrow0.08}$&$\bf{3.75(11.52\times)}$&$\bf{1.84(5.81\times)}$&$\downarrow 1.25$\\
  AT\_HE\cite{Hypersphere} &WRN-34-20&$85.14$&$62.14$&$55.33$&$54.21$&$53.74$&$43.00$&$10.69$&$\bf{53.67\downarrow0.07}$&$\bf{3.68( 11.68\times)}$&$\bf{1.81(5.91\times)}$&$\downarrow8.47$\\
  LBGAT\cite{LBGAT}$^\ddagger$   &WRN-34-20&$88.70$&$53.57$&$54.69$&$53.90$&$53.57$&$43.11$&$10.58$&$\bf{53.46\downarrow0.11}$&$\bf{3.69( 11.63\times)}$&$\bf{1.81(5.80\times)}$&$\downarrow0.11$\\
  FAT\cite{FAT}                 &WRN-34-10&$84.52$&$53.51$&$54.46$&$53.83$&$53.51$&$42.94$&$10.54$&$\bf{53.42\downarrow0.09}$&$\bf{3.68( 11.72\times)}$&$\bf{1.81(5.83\times)}$&$\downarrow0.09$\\
  Overfitting\cite{Overfitting} &WRN-34-20&$85.34$&$58.00$&$55.21$&$53.95$&$53.42$&$42.10$&$10.50$&$\bf{53.33\downarrow0.09}$&$\bf{3.66(11.50\times)}$&$\bf{1.80(5.83\times)}$&$\downarrow4.67$\\
  Self-adaptive\cite{Self-adaptive}$^\ddagger$&WRN-34-10&$83.48$&$58.03$&$54.39$&$53.62$&$53.33$&$42.10$&$10.50$&$\bf{53.20\downarrow0.13}$&$\bf{3.66(11.50\times)}$&$\bf{1.80(5.83\times)}$&$\downarrow4.83$\\
  TRADES\cite{TRADESs}$^\ddagger$&WRN-34-10&$84.92$&$56.43$&$54.02$&$53.31$&$53.08$&$42.00$&$10.40$&$\bf{53.01\downarrow0.07}$&$\bf{3.63(11.57\times)}$&$\bf{1.78(5.75\times)}$&$\downarrow3.42$\\
  LBGAT\cite{LBGAT}$^\ddagger$   &WRN-34-10&$88.22$&$52.86$&$54.37$&$53.26$&$52.86$&$41.80$&$10.30$&$\bf{52.76\downarrow0.10}$&$\bf{3.64( 11.48\times)}$&$\bf{1.79(5.79\times)}$&$\downarrow0.10$\\
  OAAT\cite{OAAT}               &ResNet-18&$80.24$&$51.06$&$51.69$&$51.28$&$51.06$&$40.54$&$10.21$&$\bf{51.02\downarrow0.04}$&$\bf{3.51( 11.53\times)}$&$\bf{1.72(5.93\times)}$&$\downarrow0.04$\\
  SAT\cite{IAR}                 &WRN-34-10&$86.84$&$50.72$&$52.95$&$51.38$&$50.72$&$40.14$&$10.01$&$\bf{50.62\downarrow0.10}$&$\bf{3.50( 11.46\times)}$&$\bf{1.72(5.81\times)}$&$\downarrow0.10$\\
  Robustness\cite{robustness}   &ResNet-50&$87.03$&$53.29$&$52.19$&$50.14$&$49.21$&$39.10$&$ 9.80$&$\bf{49.16\downarrow0.05}$&$\bf{3.42(11.43\times)}$&$\bf{1.68(5.83\times)}$&$\downarrow4.13$\\
  YOPO\cite{YOPO}               &WRN-34-10&$87.20$&$47.98$&$47.11$&$45.57$&$44.83$&$35.60$&$ 9.00$&$\bf{44.77\downarrow0.06}$&$\bf{3.09(11.52\times)}$&$\bf{1.52(5.92\times)}$&$\downarrow3.21$\\
  MMA\cite{MMA}                 &WRN-28-4&$84.36$&$47.18$&$47.78$&$42.42$&$41.51$&$33.30$&$ 8.60$&$\bf{41.27\downarrow0.24}$&$\bf{3.17( 10.50\times)}$&$\bf{1.66(5.19\times)}$&$\downarrow5.85$\\
  DNR\cite{DNR}                 &ResNet-18&$87.32$&$40.41$&$42.15$&$41.01$&$40.41$&$32.81$&$ 8.72$&$\bf{40.26\downarrow0.15}$&$\bf{2.81( 11.67\times)}$&$\bf{1.38(6.32\times)}$&$\downarrow5.93$\\
  CNL\cite{level-sets}$^\ddagger$&ResNet-18&$81.30$&$79.67$&$40.26$&$40.23$&$40.22$&$32.70$&$8.70$&$\bf{39.83\downarrow0.39}$&$\bf{2.74(11.93\times)}$&$\bf{1.34(6.49\times)}$&$\downarrow39.84$\\
  Feature Scatter\cite{Feature-scatter} &WRN-28-10&$89.98$&$60.60$&$54.63$&$42.91$&$36.62$&$30.00$&$ 8.20$&$\bf{36.31\downarrow0.33}$&$\bf{11.02( 2.72\times)}$&$\bf{5.44(1.51\times)}$&$\downarrow24.33$\\
  Interpolation\cite{Interpolation}     &WRN-28-10&$90.25$&$68.70$&$66.72$&$49.35$&$36.45$&$30.00$&$ 8.50$&$\bf{36.21\downarrow0.24}$&$\bf{11.21( 2.64\times)}$&$\bf{5.52(1.54\times)}$&$\downarrow32.32$\\
  Sensible\cite{Sensible}               &WRN-34-10&$91.51$&$57.23$&$56.04$&$43.15$&$34.22$&$28.20$&$ 7.80$&$\bf{34.00\downarrow0.22}$&$\bf{10.66( 2.65\times)}$&$\bf{5.25(1.49\times)}$&$\downarrow23.23$\\
  Regularization\cite{Regualr}  &ResNet-18&$90.84$&$77.68$&$52.77$&$19.73$&$1.35$&$ 3.10$&$ 2.30$&$\bf{ 0.89\downarrow0.46}$&$\bf{2.24( 1.38\times)}$&$\bf{1.09(2.11\times)}$&$\downarrow76.79$\\\midrule\midrule
    \multirow{2}{*}{\textbf{\makecell[c]{CIFAR-100 \\Defense Method}}}&\multirow{2}{*}{\textbf{Model}}&\textbf{Clean}&\textbf{Nominal}&\textbf{PGD}&\textbf{ODI}&\multicolumn{3}{c}{\textbf{AA}}&\multicolumn{3}{c}{\textbf{A$^{3}$}}&$\bf{\Delta}$\\\cmidrule(lr){3-3}\cmidrule(lr){4-4}\cmidrule(lr){5-5}\cmidrule(lr){6-6}\cmidrule(lr){7-9}\cmidrule(lr){10-12}\cmidrule(l){13-13}
                                   &    & acc & acc & acc   & acc & acc &$\rightarrow$&$\leftarrow$& acc &$\rightarrow$&$\leftarrow$& acc    \\\midrule
  ULAT\cite{ULAT}$^\dagger$     &WRN-70-16&$69.15$&$36.88$&$38.64$&$37.41$&$36.88$&$29.84$&$7.42$&$\bf{36.86\downarrow0.02}$&$\bf{2.56(11.64\times)}$&$\bf{1.25(5.92\times)}$&$\downarrow0.02$\\
  Fixing Data\cite{fix_data}      &WRN-70-16&$63.56$&$34.64$&$35.95$&$34.98$&$34.64$&$28.02$&$6.96$&$\bf{34.55\downarrow0.09}$&$\bf{2.38(11.76\times)}$&$\bf{1.16(6.00\times)}$&$\downarrow0.04$\\
  Fixing Data\cite{fix_data}      &WRN-28-10&$62.41$&$32.06$&$33.39$&$32.36$&$32.06$&$25.53$&$6.48$&$\bf{32.00\downarrow0.06}$&$\bf{2.24(11.38\times)}$&$\bf{1.10(5.90\times)}$&$\downarrow0.06$\\
  OAAT\cite{OAAT}               &WRN-34-10&$65.73$&$30.35$&$31.62$&$30.93$&$30.35$&$24.34$&$6.11$&$\bf{30.31\downarrow0.04}$&$\bf{2.18(11.14\times)}$&$\bf{1.07(5.70\times)}$&$\downarrow0.04$\\
  LBGAT\cite{LBGAT}$^\ddagger$             &WRN-34-20&$62.55$&$30.20$&$31.65$&$30.49$&$30.20$&$23.97$&$6.10$&$\bf{30.12\downarrow0.08}$&$\bf{2.16( 11.11\times)}$&$\bf{1.05(5.80\times)}$&$\downarrow0.08$\\
  ULAT\cite{ULAT}     &WRN-70-16&$60.86$&$30.03$&$31.03$&$30.41$&$30.03$&$23.93$&$6.09$&$\bf{29.99\downarrow0.04}$&$\bf{2.13(11.23\times)}$&$\bf{1.04(5.86\times)}$&$\downarrow0.04$\\
  LBGAT\cite{LBGAT}$^\ddagger$             &WRN-34-10&$60.64$&$29.33$&$30.56$&$29.63$&$29.33$&$23.21$&$5.94$&$\bf{29.18\downarrow0.15}$&$\bf{2.11( 11.00\times)}$&$\bf{1.03(5.77\times)}$&$\downarrow0.15$\\
  AWP\cite{AWP}       &WRN-34-10&$60.38$&$28.86$&$30.70$&$29.45$&$28.86$&$23.01$&$5.84$&$\bf{28.78\downarrow0.08}$&$\bf{2.10(10.96\times)}$&$\bf{1.02(5.72\times)}$&$\downarrow0.08$\\
  Pre-training\cite{Pre-training} &WRN-28-10&$59.23$&$28.42$&$30.56$&$29.13$&$28.42$&$22.74$&$5.73$&$\bf{28.31\downarrow0.11}$&$\bf{2.08(10.93\times)}$&$\bf{1.02(5.61\times)}$&$\downarrow0.11$\\
  OAAT\cite{OAAT}               &ResNet18&$62.02$&$27.14$&$27.90$&$27.47$&$27.14$&$21.74$&$5.61$&$\bf{27.09\downarrow0.05}$&$\bf{2.34(9.29\times)}$&$\bf{1.15(4.88\times)}$&$\downarrow0.05$\\
  SAT\cite{IAR}                 &WRN-34-10&$62.82$&$24.57$&$26.69$&$25.43$&$24.57$&$19.70$&$5.10$&$\bf{24.51\downarrow0.06}$&$\bf{1.90(10.36\times)}$&$\bf{0.93(5.48\times)}$&$\downarrow0.06$\\
  Overfitting\cite{Overfitting} &PAResNet-18&$53.83$&$18.95$&$20.15$&$19.39$&$18.95$&$15.28$&$4.00$&$\bf{18.90\downarrow0.05}$&$\bf{1.64(9.32\times)}$&$\bf{0.80(5.00\times)}$&$\downarrow0.05$\\\bottomrule
  \end{tabular}}
  \caption{Comparison of robust accuracy (\%) under the attack of A$^{3}$, PGD, ODI, and AutoAttack(AA) across various defense strategies. 
    The ``acc" column shows the robust accuracies of different models.
    The ``Nominal" column shows the robust accuracies reported by defense models.
    The ``$\Delta$" column shows the difference between the robust accuracies of ``Nominal" and A$^{3}$. 
    The ``$\rightarrow$" column shows the iteration number of forward propagation (million), while the ``$\leftarrow$" column shows the iteration number of backward propagation (million). Models marked with $\dagger$ were additionally trained with unlabeled datasets. We used $\epsilon$ = 8/255 except for models marked with $\ddagger$ , which used $\epsilon$ = 0.031 as originally reported by the authors.
    Notably, the ``\textbf{acc}" column of A$^3$ shows the difference between the robust accuracies of AA and A$^3$, the ``$\leftarrow$" and ``$\rightarrow$" columns of A$^{3}$ show the speedup factors of A$^3$ relative to AA.}
    \label{table.main}
  \end{table*}

\subsection{Comparisons with State-of-the-Art Attacks}
\label{sec:compare with sota}
To comprehensively validate the efficiency and reliability of our method, we compare AA, PGD, ODI with our A$^3$ on nearly 50 defense models.
The evaluation results are shown in \cref{table.main}.

\noindent{\textbf{Setup.}} Following the setup of ODI and PGD, 100 iterations are allocated for each image (4 restarts, 25 iterations for attacks at each restart), $N_{odi}=2$. 
For AA, the standard version\footnote{\url{https://github.com/fra31/auto-attack}} is adopted. For our A$^3$, initial number of iterations for attacking $\gamma$ is set to 25, and the iteration increment $\nu$ is 5. The number of iteration for initialization $N=7$.
When the number of iterations reaches 50, we keep it unchanged to save the budget number of iterations. 
Initial discarding rate $\phi=0$ and discarding increment $\iota=0.1$, when the discarding rate reaches to 0.9, we gradually increase it to 0.97 at an interval of $\iota = 0.035$.

\noindent{\textbf{Results.}}
As can be seen in \cref{table.main}, A$^3$ uses the same parameters for all defense models and achieves lower robust accuracy than AA in all cases, downgrading \textbf{acc} by 0.1\% on average. 
Besides, our method achieves a faster evaluation, on average, $10.4\times$ speed up for forward propagation, and $5.4\times$ for backward propagation.
In general, the nature of parameter-free, reliability, and efficiency enables A$^3$ to be a practical method for robustness evaluation. 
For the results on other datasets, network architectures and metrics (~\ie, {$\ell_2$}-norm), please refer to the Appendix. C.
\subsection{Ablation Study}
\noindent{\textbf{Efficacy of adaptive direction initialization.}} To evaluate the effectiveness of ADI, we design another variant called Reverse Adaptive Direction Initialization (R-ADI) which adopts the reverse direction of the adaptive direction. 
For all methods, we allocate 150 iterations for each image (5 restarts and 30 iterations at each restart.), and $N$=10. 

The comparison results among R-ADI, ODI, and ADI are reported in \cref{table:two_modules}.
As can be seen, compared to ODI, our ADI achieves better attack performance in all cases. The result indicates the initial direction does affect the performance. 
Compared with uniformly generating initial direction, our ADI can generate model-specific initial direction and help to obtain better performance. 
In general, R-ADI achieves the worst performance in all cases. 
One possible reason is that R-ADI chooses a bad initial direction, which hinders the performance.


\noindent{\textbf{Efficacy of online statistics-based discarding strategy.}} To verify the effect of OSD,  we compare the robust accuracy curves  of the defense models under the attack of ADI, ADI+OSD (A$^{3}$), and AA. For our  methods ADI and ADI+OSD, the setup is the same as in \cref{sec:compare with sota}.

\begin{figure*}[htb]
\centering
\begin{minipage}[t]{0.24\linewidth}
\includegraphics[width=4.2cm]{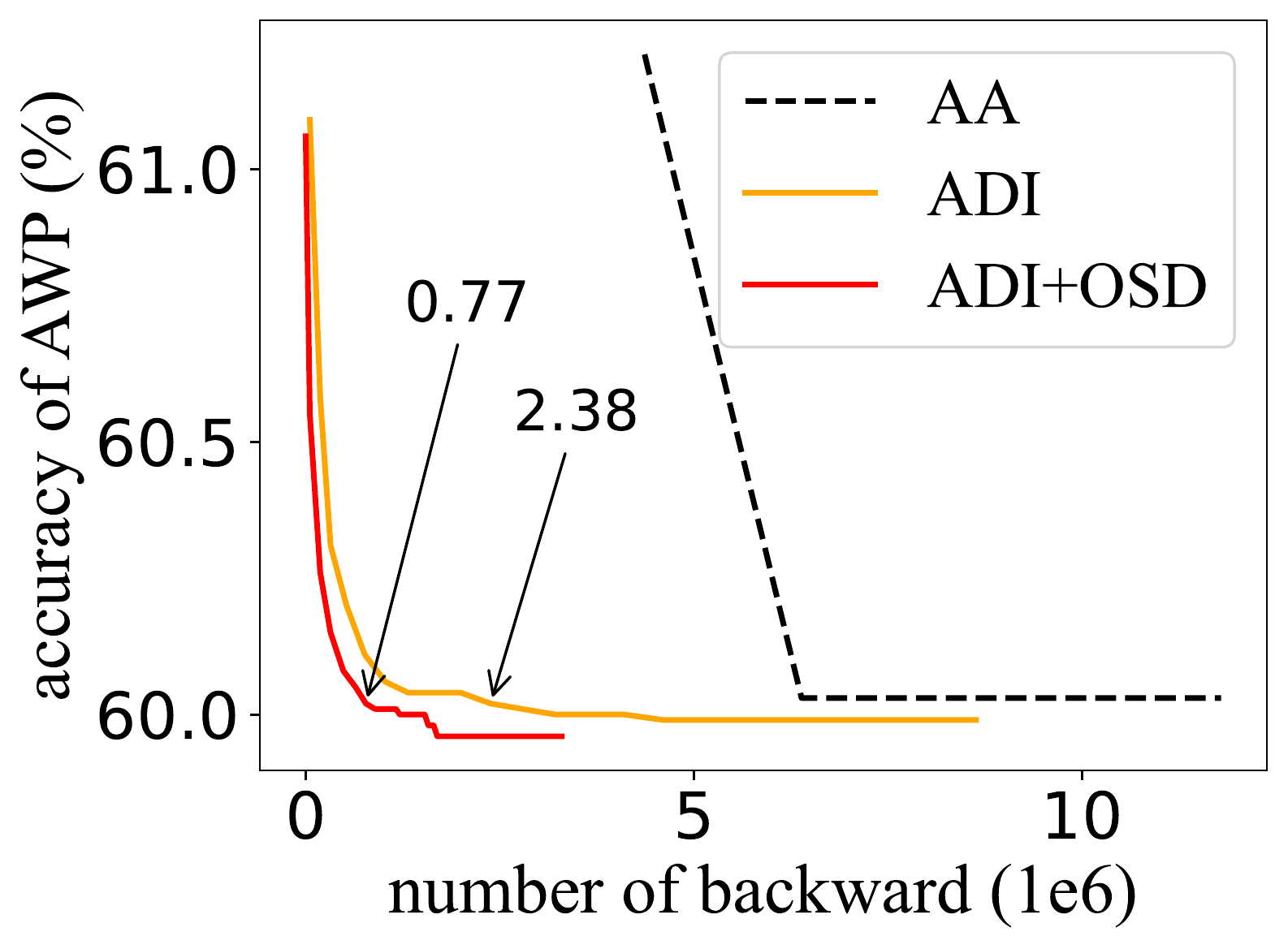}
\end{minipage}
\begin{minipage}[t]{0.23\linewidth}
\includegraphics[width=4.0cm]{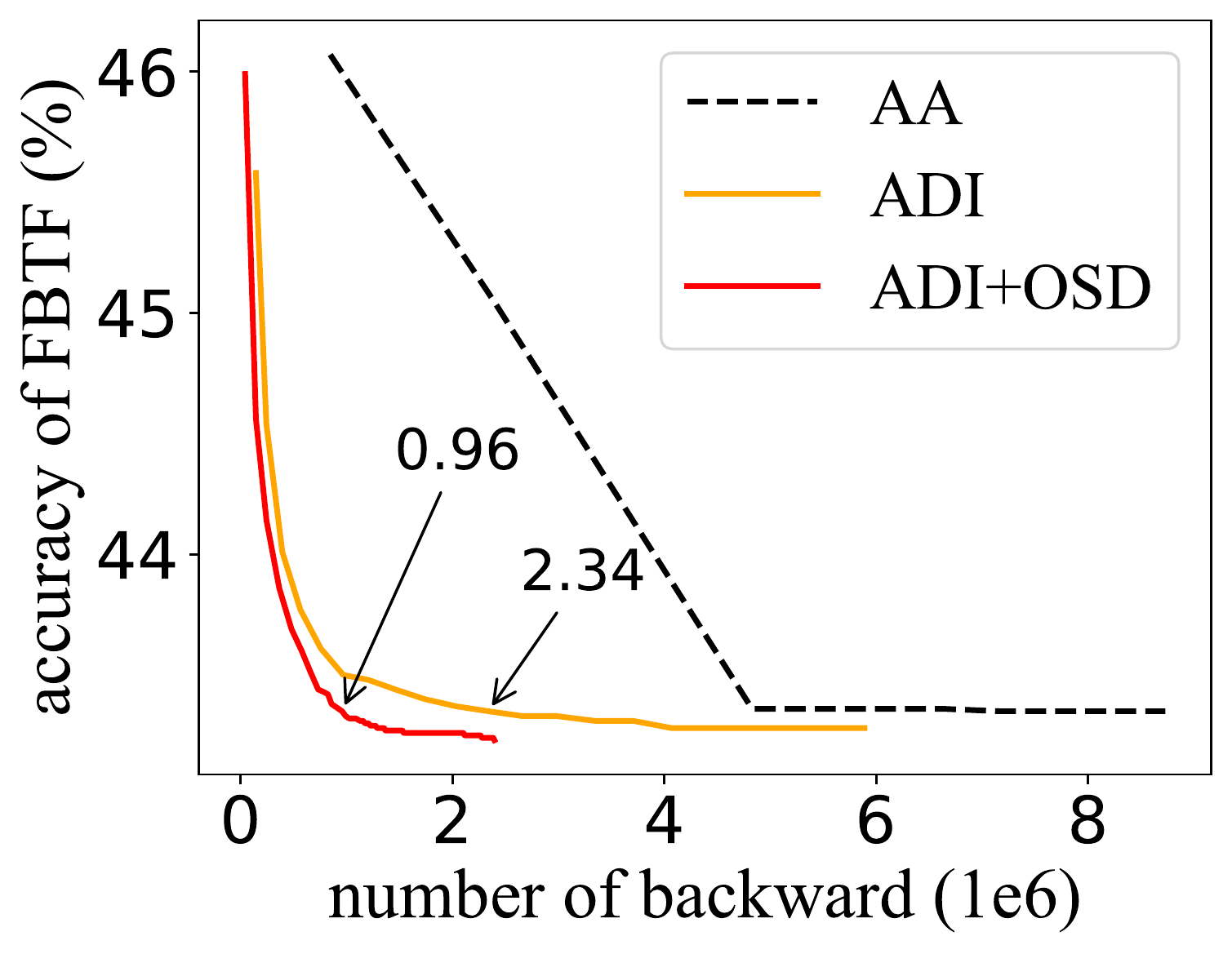}
\end{minipage}
\begin{minipage}[t]{0.24\linewidth}
\includegraphics[width=4.2cm]{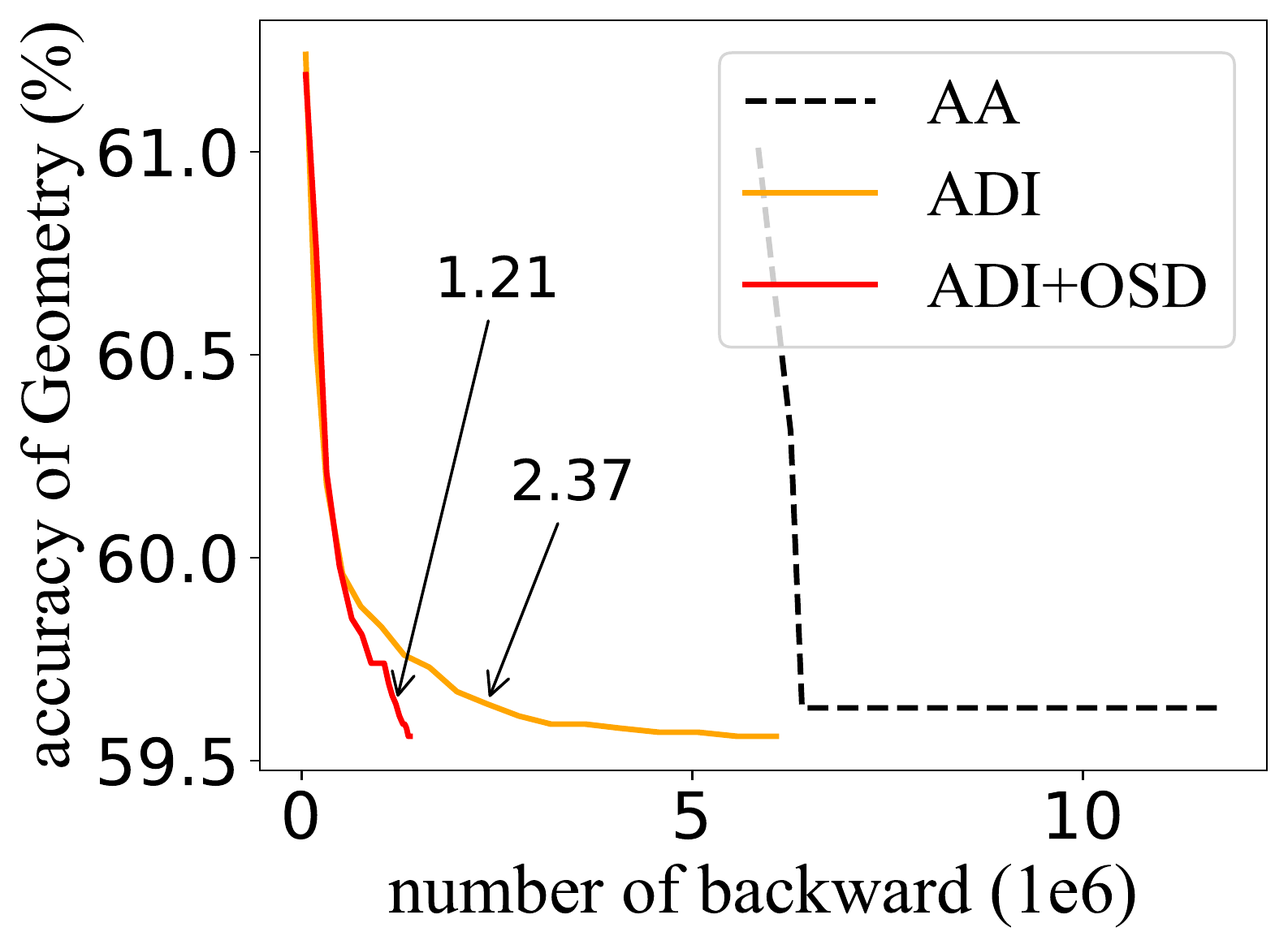}
\end{minipage}
\begin{minipage}[t]{0.23\linewidth}
\includegraphics[width=4.0cm]{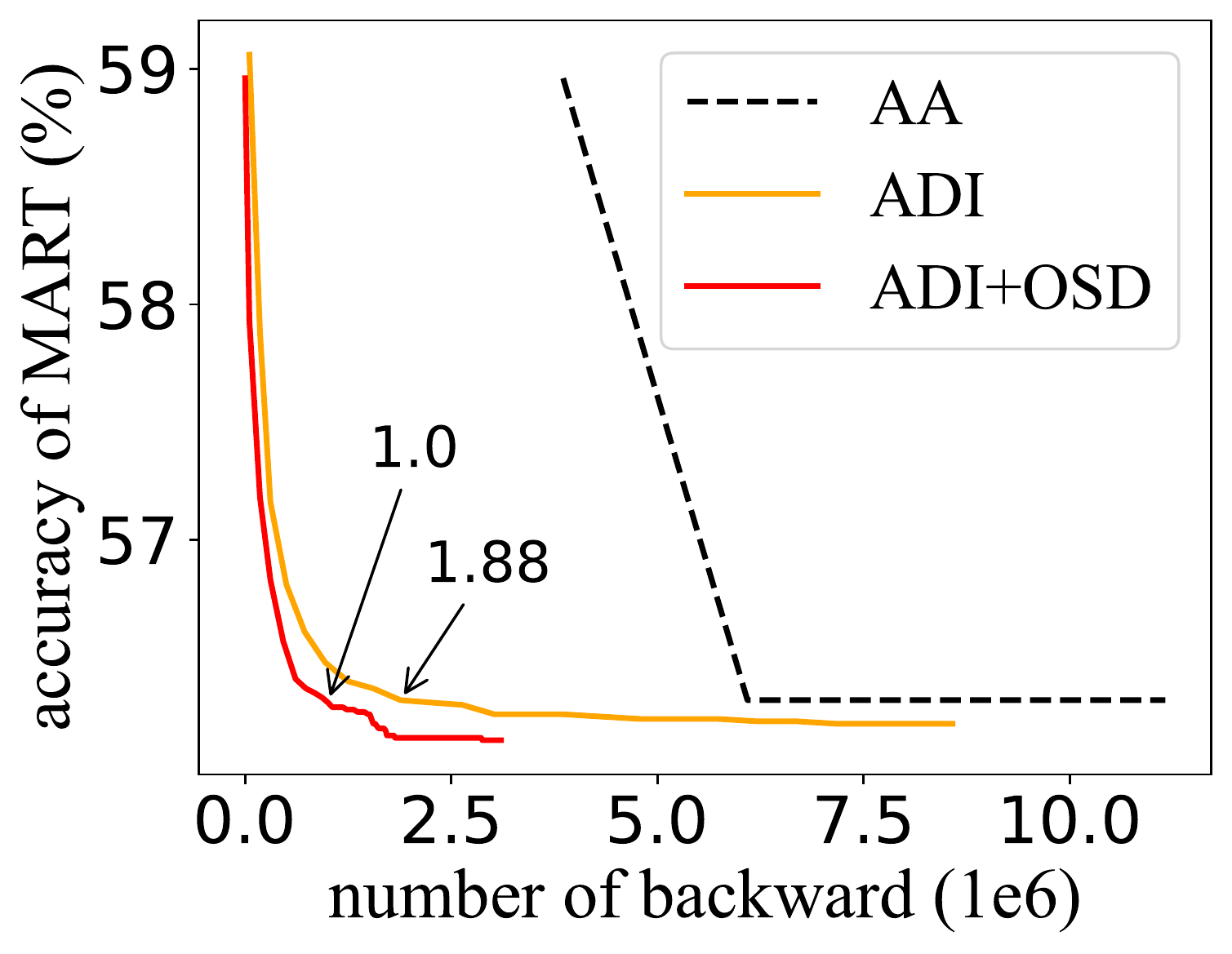}
\end{minipage}
\begin{minipage}[t]{0.24\linewidth}
\includegraphics[width=4.2cm]{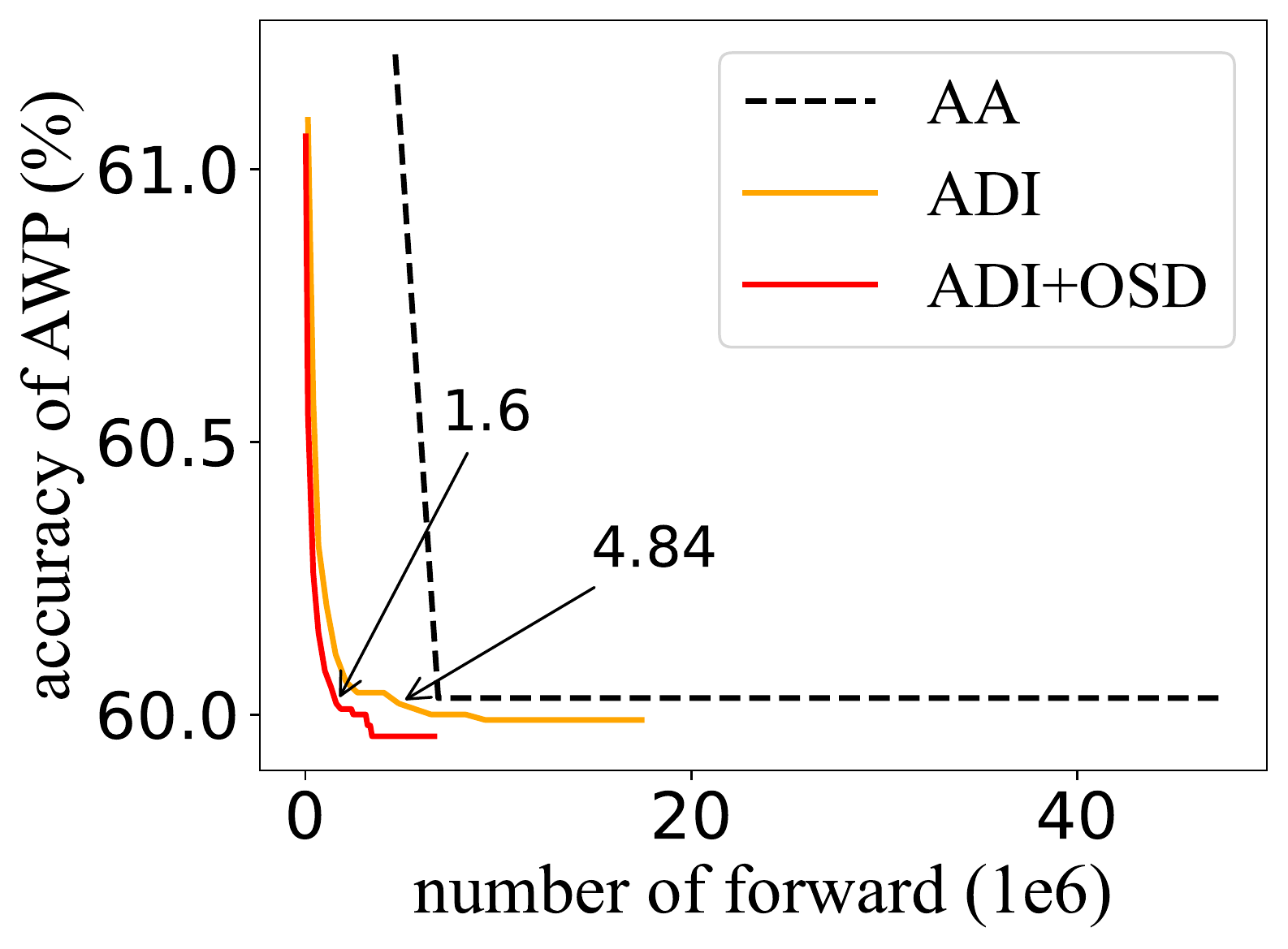}
\end{minipage}
\begin{minipage}[t]{0.23\linewidth}
\includegraphics[width=4cm]{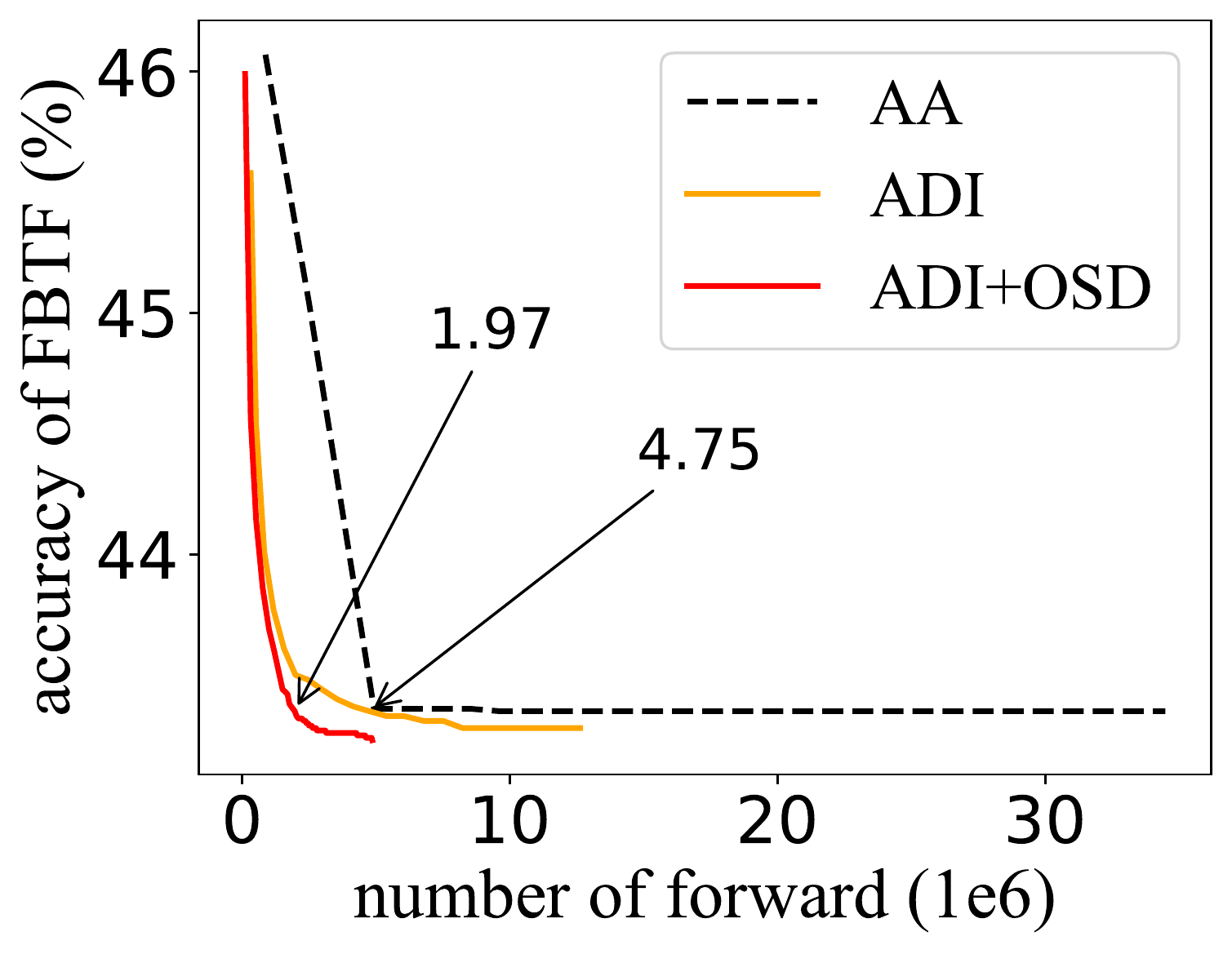}
\end{minipage}
\begin{minipage}[t]{0.24\linewidth}
\includegraphics[width=4.2cm]{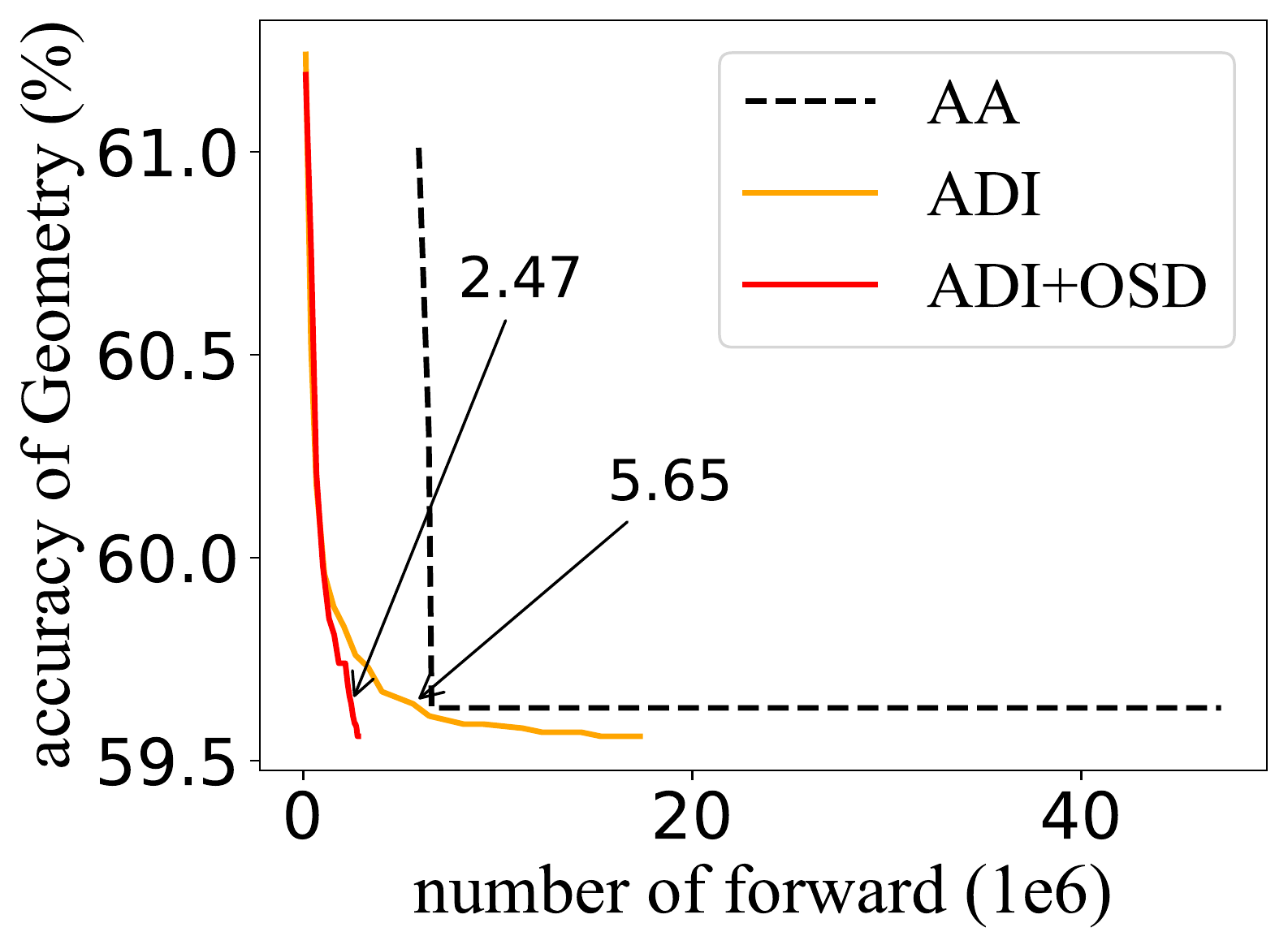}
\end{minipage}
\begin{minipage}[t]{0.23\linewidth}
\includegraphics[width=4.0cm]{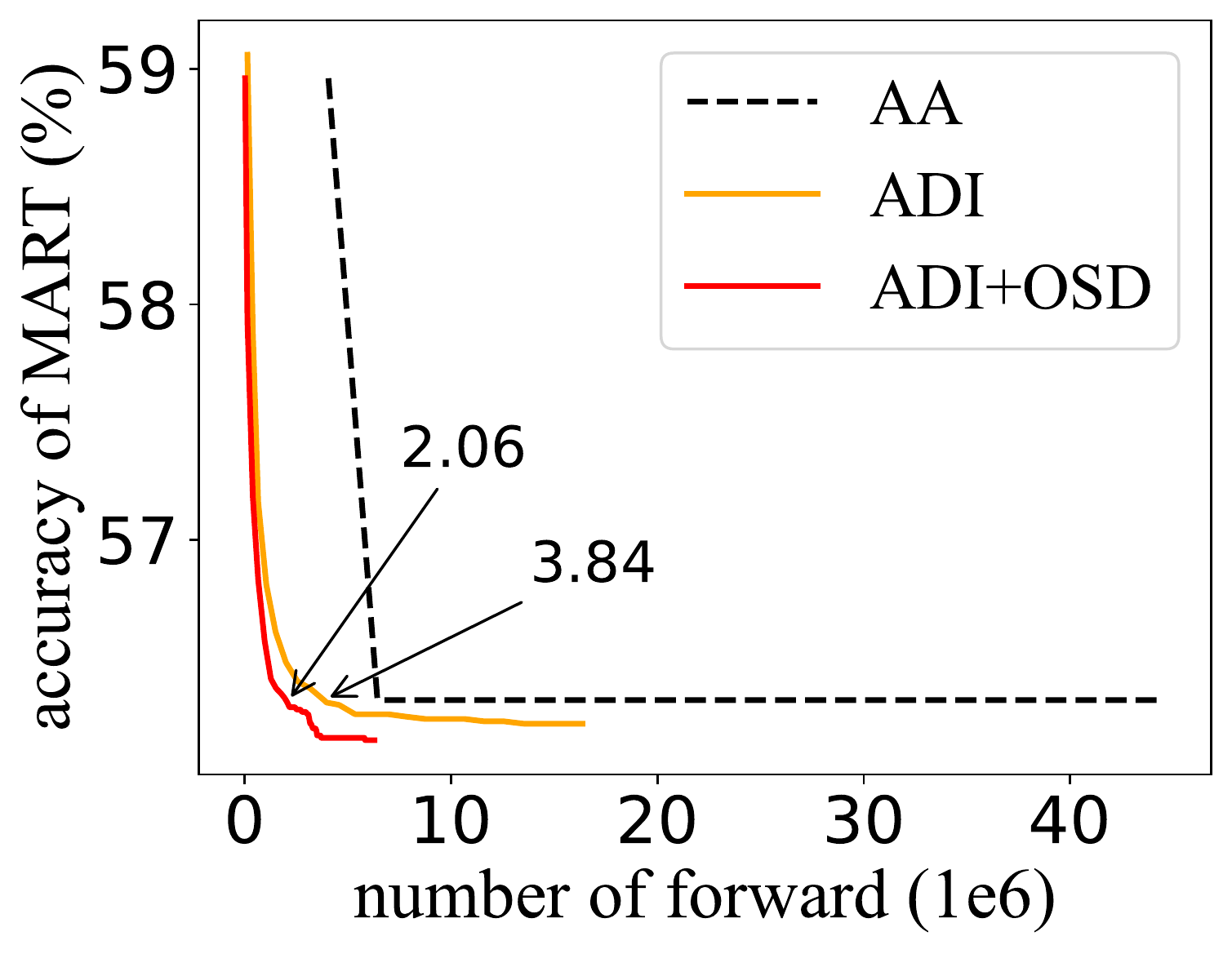}
\end{minipage}
\caption{Comparisons of performance of ADI+OSD, ADI and AA on 4 defenders. For each defender, we separately record the number of forward propagation and back propagation by columns. The horizontal axes show the number of back propagation (top) and forward propagation (bottom). The vertical axes show the percentage of remaining unsuccessful examples. 
The iteration numbers needed for ADI+OSD and ADI to defeat AA are also marked.}
\vspace{-0.7em}
\label{fig:OSD}
\end{figure*}

Note that ADI+OSD denotes the Online Statistics-based Discarding strategy is applied on ADI. The result is shown in the \cref{fig:OSD}.
As can be observed, it costs more iterations for AA to achieve the same robust accuracy of ADI and ADI+OSD in all cases. 
Meanwhile, to achieve higher attack performance, ADI and AA require a large number of additional iterations, but ADI+OSD requires fewer iterations.
The curves reveal the efficiency of our ADI+OSD, especially for reliable attacks. 

\noindent{\textbf{Efficacy on other robustness evaluation methods.}} In this section, we study the effect of integrating ADI and OSD into different robustness evaluation methods.
For all methods, we allocate 500 iterations for each image (5 restarts and 100 iterations for attacks at each restart), for ODI method, $N_{odi}$=10. For MT-PGD, the number of multiple targets is 3. For our A$^3$, $N=10$ . The other experimental settings are the same as in \cref{sec:compare with sota}.

\cref{tab:other_methods} shows the robust accuracy~(\%) of defense models under the attack of different attack methods~\cite{ifgsm,pgd,odi,mt-pgd} integrated with our modules (ADI and OSD).  
It shows that ADI and OSD can be integrated into multiple attack methods and effectively improve their performance.

     

\subsection{Result of Competition}
With our proposed A$^3$, we participated in the CVPR 2021 White-box Adversarial Attacks on Defense Models competition launched by Alibaba Group and Tsinghua University. 
To evaluate performance fairly, all codes were tested with the official adversarial robustness evaluation platform ARES. 
The competition has three stages. Stage 1 and stage 2 employed 13 and 2 defense models trained on CIFAR-10 and ImageNet datasets, respectively. 
The $\ell_\infty$ constraint is set to $8/255$, $4/255$ on CIFAR-10 and  ImageNet, respectively. 
Methods for all participants tested on CIFAR-10 dataset and $1,000$ random images of the ImageNet validation set to evaluate the final score. The evaluation score is defined as the average misclassification rate.
Finally, we obtained $51.10\%$ score in the final stage and achieved first place among 1681 teams.
\begin{table}[!t]
  \centering
  \label{table.direction}
  \resizebox{0.8\columnwidth}{!}{
  \begin{tabular}{@{}c|ccc@{}}
  \toprule
  \textbf{Defense method}    &\textbf{R-ADI}     &\textbf{ODI}      &\textbf{ADI}  \\ \midrule
   Geometry &$60.00\uparrow0.09$&$59.91$&$\bf{59.73\downarrow 0.18}$   \\
   RST              &$59.81\uparrow0.01$&$59.80$& $\bf{59.63\downarrow0.17}$   \\
   Proxy            &$59.65\uparrow0.25$&$59.40$& $\bf{59.22\downarrow0.18}$   \\
   HYDRA            &$57.47\uparrow0.11$&$57.36$& $\bf{57.26\downarrow0.10}$   \\
   MART             &$56.86\uparrow0.28$&$56.58$& $\bf{56.48\downarrow0.10}$   \\
   Pre-training     &$55.48\uparrow0.37$&$55.11$& $\bf{54.96\downarrow0.15}$   \\
   Self-adaptive    &$53.59\uparrow0.11$&$53.48$& $\bf{53.33\downarrow0.15}$   \\
   Feature Scatter            &$40.90\uparrow2.22$&$38.68$& $\bf{38.29\downarrow0.39}$   \\
   Interpolation              &$55.47\uparrow12.95$&$42.52$& $\bf{40.01\downarrow2.51}$  \\
   Sensible                   &$41.14\uparrow3.19$&$37.95$& $\bf{37.19\downarrow0.76}$   \\
   Regularization             &$24.82\uparrow14.83$&$ 9.99$& $\bf{7.75\downarrow2.24}$   \\ \bottomrule
  \end{tabular}
  }
  \caption{ Comparisons of robust accuracy (\%) under attack of ADI, na{\"i}ve ODI and R-ADI across various defense strategies. R-ADI obtains the worst results because it chose the worst starting points.}
  \label{table:two_modules}
\end{table}

\section{Conclusion}
In this paper, we find that model-agnostic initial directions drawn from uniform distributions result in unsatisfactory robust accuracy, and na{\" i}ve iterative strategy leads to unreliable attacks. We propose a novel approach called Adaptive Auto Attack~(A$^3$) to address these issues, which adopts adaptive direction initialization and an online statistics-based discarding strategy to achieve efficient and reliable robustness evaluation. Extensive experiments demonstrate the effectiveness of A$^3$. Particularly, we achieve lower robust accuracy in all cases by consuming much fewer iterations than existing models,~\eg, $1/10$ on average (10$\times$ speed up). We won first place out of 1,681 teams in CVPR 2021 White-box Adversarial Attacks on Defense Models competitions with this method.

\begin{table}[t]
\centering
\resizebox{0.48\textwidth}{!}{
\begin{tabular}{l|cccc}
\hline
Attack method   & RLPE & Self-adaptive & OAAT  & Feature Scatter \\ \hline
I-FGSM         & $60.77$ & $54.38$         & $51.74$ & $55.00$           \\
I-FGSM+ADI      & $59.66\downarrow 1.12$ & $53.36\downarrow 1.02$   & $51.09\downarrow 0.65$      &  $38.24\downarrow 16.76$   \\
I-FGSM+ADI+OSD   & $\textbf{59.50}\downarrow \textbf{1.28}$ & $\textbf{53.24}\downarrow \textbf{1.14}$ & $\textbf{51.01}\downarrow \textbf{0.73}$ &  $\textbf{36.70}\downarrow \textbf{18.30}$ \\ \hline
PGD            & $60.65$ & $54.29$         & $51.5$ & $52.98$           \\
PGD+ADI         & $59.61\downarrow 1.04$ & $53.37\downarrow 0.92$            &$51.08\downarrow 0.42$       &  $38.08\downarrow 14.90$               \\
PGD+ADI+OSD  & $\textbf{59.51}\downarrow \textbf{1.14}$ & $\textbf{53.23}\downarrow \textbf{1.06}$& $\textbf{51.04}\downarrow \textbf{0.46}$ & $\textbf{36.72}\downarrow \textbf{16.26}$ \\ \hline
MT-PGD         & $60.51$ & $54.08$         & $51.59$ & $50.92$           \\
MT-PGD+ADI     & $60.11\downarrow 0.40$ & $53.69\downarrow 0.39$            & $51.50\downarrow 0.09$      &  $38.41\downarrow 12.51$               \\
MT-PGD+ADI+OSD  & $\textbf{59.71}\downarrow \textbf{0.80}$ & $\textbf{53.34}\downarrow \textbf{0.74}$ &  $\textbf{51.18}\downarrow \textbf{0.41}$ &  $\textbf{36.74}\downarrow \textbf{14.18}$ \\ \hline
ODI-PGD       & $59.74$ & $53.49$         & $51.13$ & $38.56$           \\
ODI-PGD+ADI   & $59.65\downarrow 0.09$ & $53.34\downarrow 0.15$            & $51.09\downarrow 0.04$      & $37.92\downarrow 0.64$                \\
ODI-PGD+ADI+OSD  & $\textbf{59.51}\downarrow \textbf{0.23}$ & $\textbf{53.26}\downarrow \textbf{0.23}$  & $\textbf{51.00}\downarrow \textbf{0.13}$ &  $\textbf{36.75}\downarrow \textbf{1.17}$ \\ \hline
\end{tabular}}
\caption{Effectiveness of the two modules ADI and OSD on 4 different attack methods.}
\vspace{-1.0em}
\label{tab:other_methods}
\end{table}

\section{Broader Impacts}
Performing reliable robustness evaluation helps  distinguish good from bad defenses to resist against the widespread adversarial examples. Our research proposes a practical robustness evaluation method. On the positive side, our method enables us to identify advanced defenses to defend against adaptive attacks, preventing critical safety systems from crashing. On the negative side, malicious users can exploit our method to attack the system, raising a security risk. 
We will continue to expand the scope of our evaluation for advanced defenses in the future.

\section*{Acknowledgements}
This work is supported by the National Natural Science Foundation of China (Grant No. 62020106008, No. 62122018, No. 61772116, No. 61872064), Sichuan Science and Technology Program (Grant No.2019JDTD0005).
\newpage
{\small
\bibliographystyle{ieee_fullname}
\bibliography{egbib}
}

\newpage

\appendix

\section{Introduction}
Due to the page limitation of the paper, we further illustrate our method in this supplementary material, which
contains the following sections: 1). Detailed quantitative results of the diversified direction $\vect w_{\vect d}$; 2). The results of the proposed A$^{3}$ attack across various defense strategies, datasets, network architectures and metrics.


\section{Detailed quantitative results of the diversified direction $\vect{w_d}$ }
In section 3.2 of the main paper, to illustrate that random sampling is sub-optimal, we use ODI~\cite{odi} to attack 11 defense models, and only give the mean values of $\vect{w_d}$ at the $\hat{y}$-th (the misclassification label), and the $y$-th (the ground truth).

In order to observe the detailed quantitative results of the diversified direction  $\vect w_{\vect d}$.
In this section, we use ODI~\cite{odi} to attack 12 defense models,  including  AWP~\cite{AWP}, Proxy~\cite{Proxy}, Fast~\cite{fbtf}, Feature Scatter~\cite{Feature-scatter}, Geometry~\cite{Geometry}, HYDRA~\cite{HYDRA}, Hypersphere~\cite{Hypersphere}, Interpolation~\cite{Interpolation}, Regular~\cite{Regualr}, MART~\cite{MARTs}, MMA~\cite{MMA} and Pre-training~\cite{Pre-training}. The experiment settings are the same as the section 3.2 of the main paper. 
The CIFAR-10 dataset is used in this experiment, there are a total of 10 categories, with 9 error categories and one ground truth.

\begin{figure*}[t]
\centering
\begin{minipage}[t]{0.45\linewidth}
\includegraphics[width=7cm]{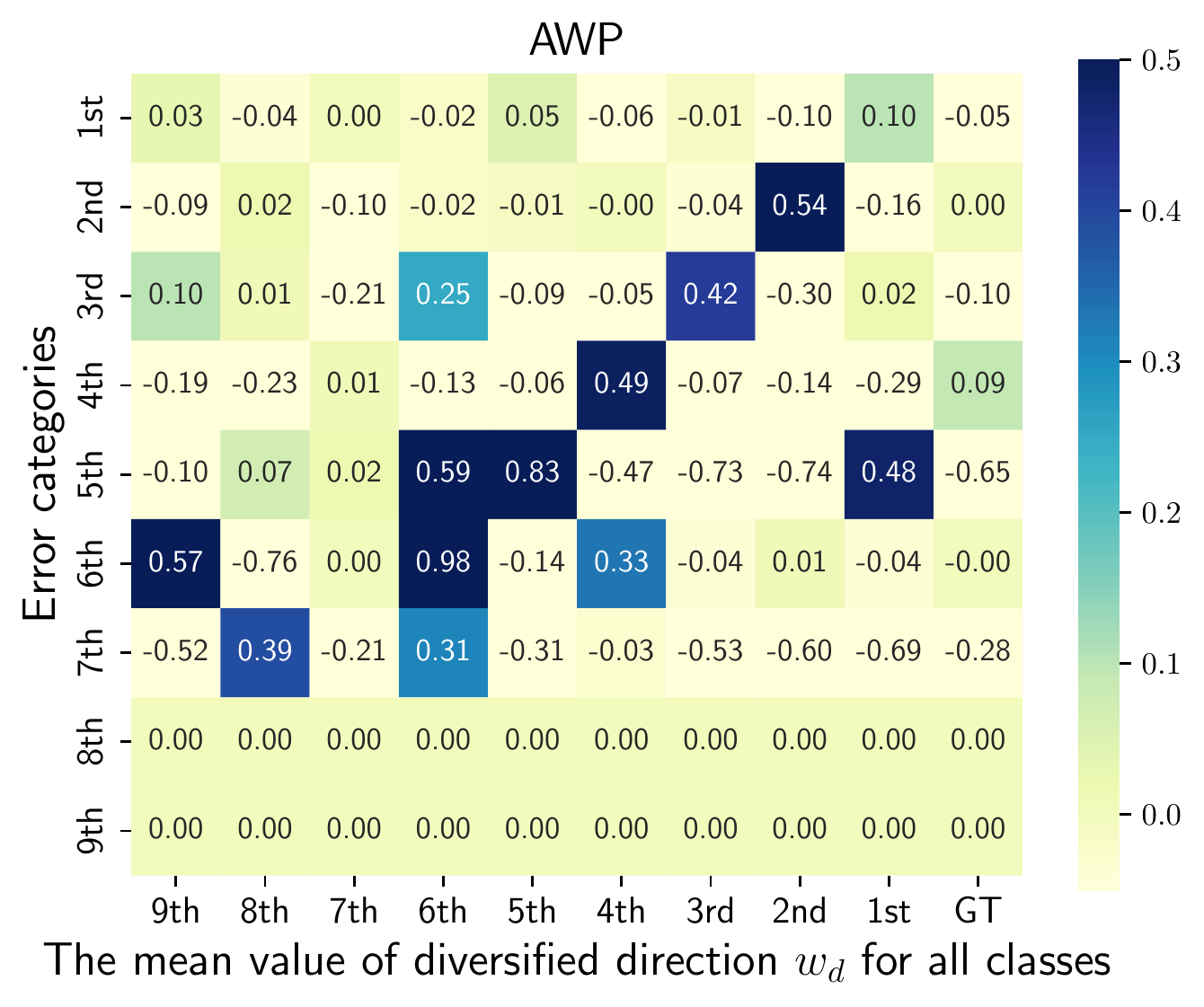}
\end{minipage}
\begin{minipage}[t]{0.45\linewidth}
\includegraphics[width=7.0cm]{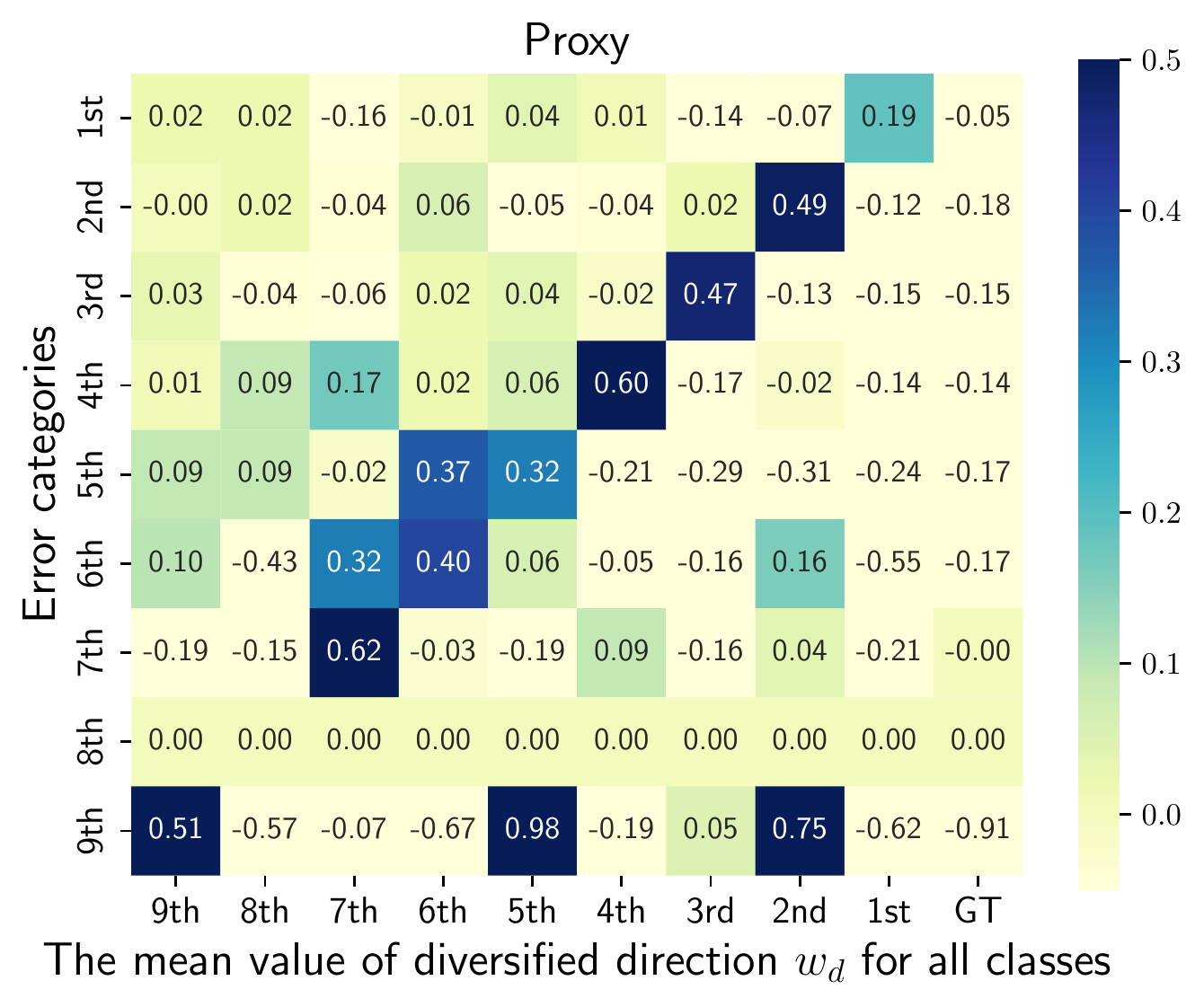}
\end{minipage}
\begin{minipage}[t]{0.45\linewidth}
\includegraphics[width=7.0cm]{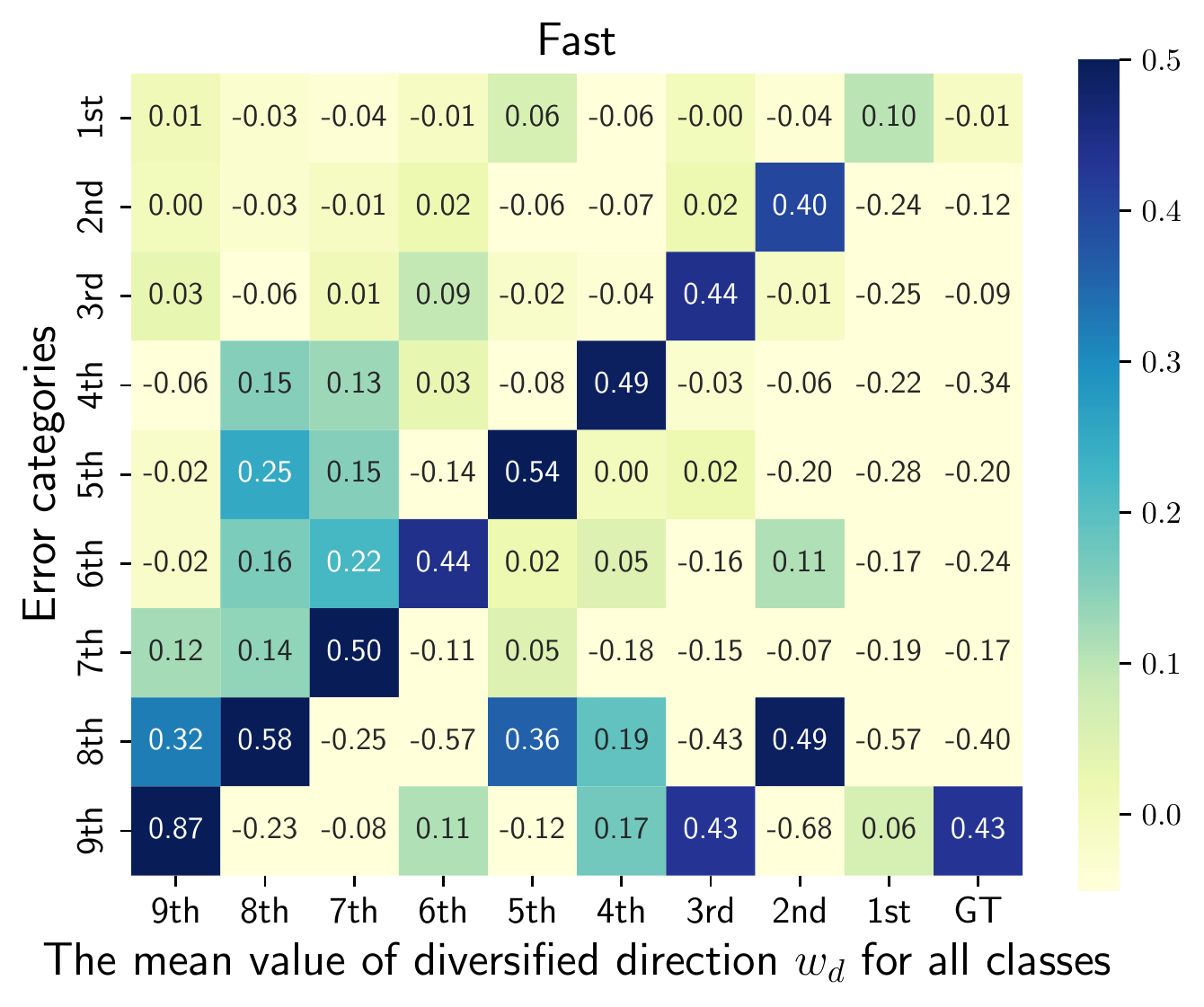}
\end{minipage}
\begin{minipage}[t]{0.45\linewidth}
\includegraphics[width=7.0cm]{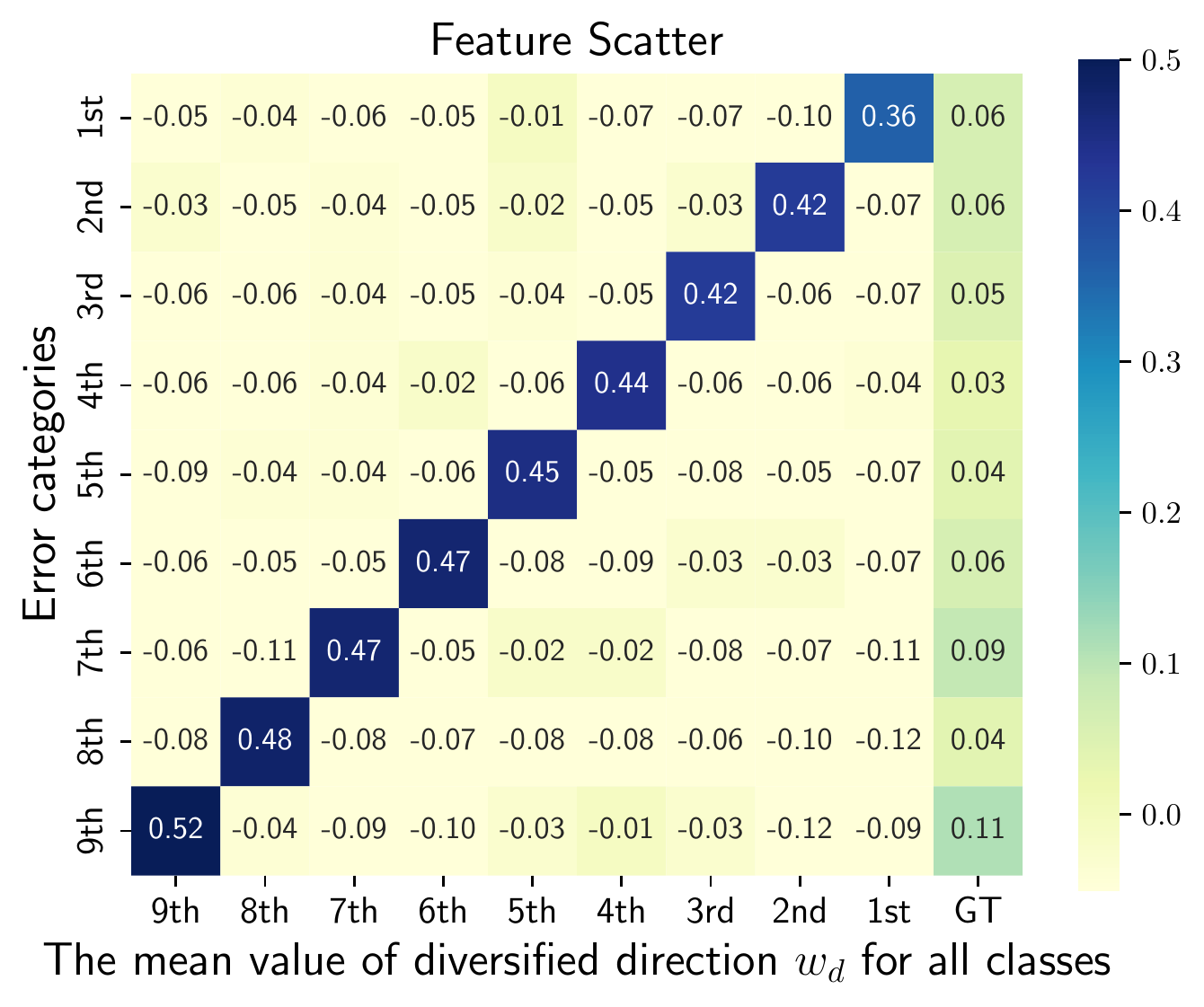}
\end{minipage}
\begin{minipage}[t]{0.45\linewidth}
\includegraphics[width=7.0cm]{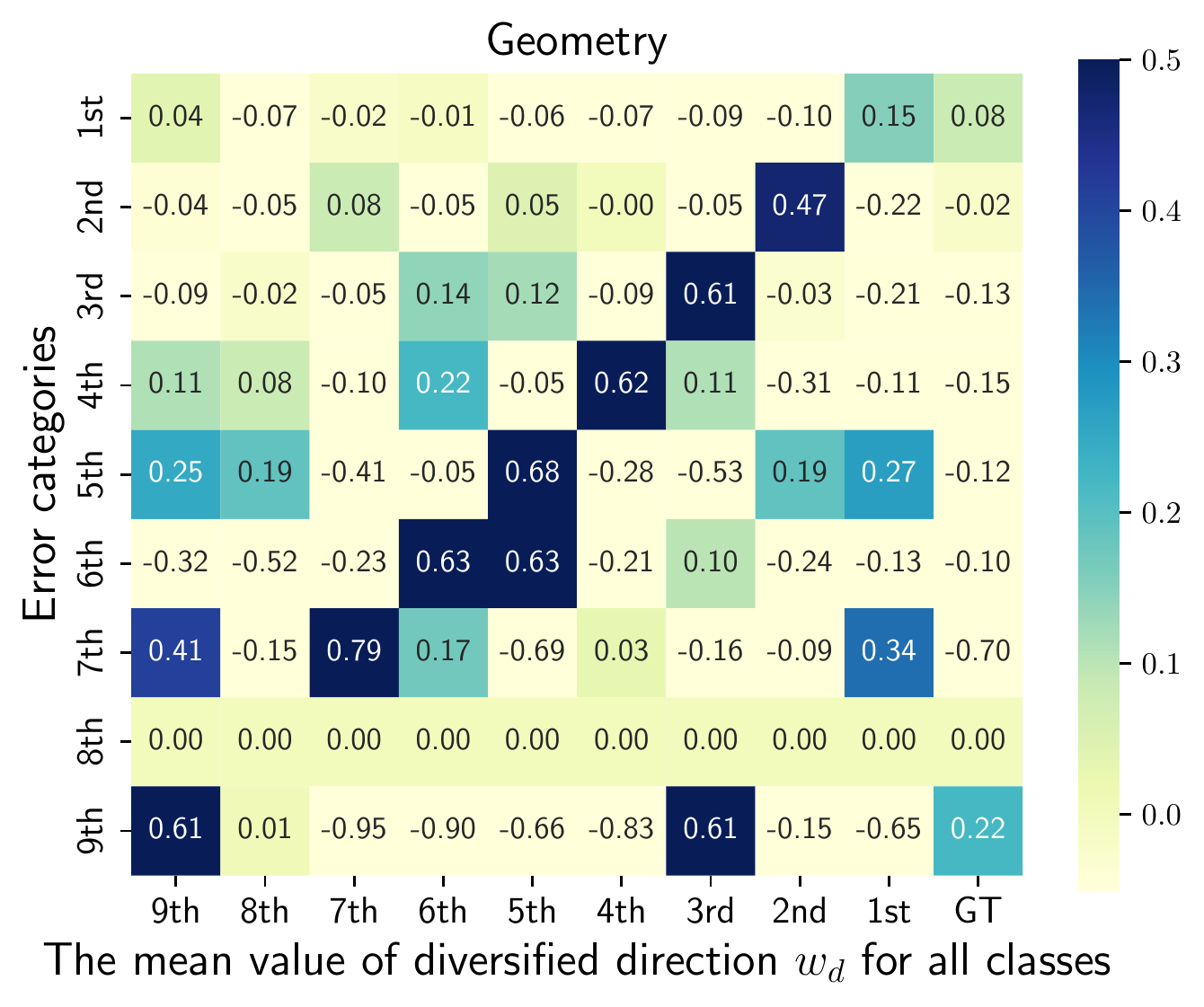}
\end{minipage}
\begin{minipage}[t]{0.45\linewidth}
\includegraphics[width=7cm]{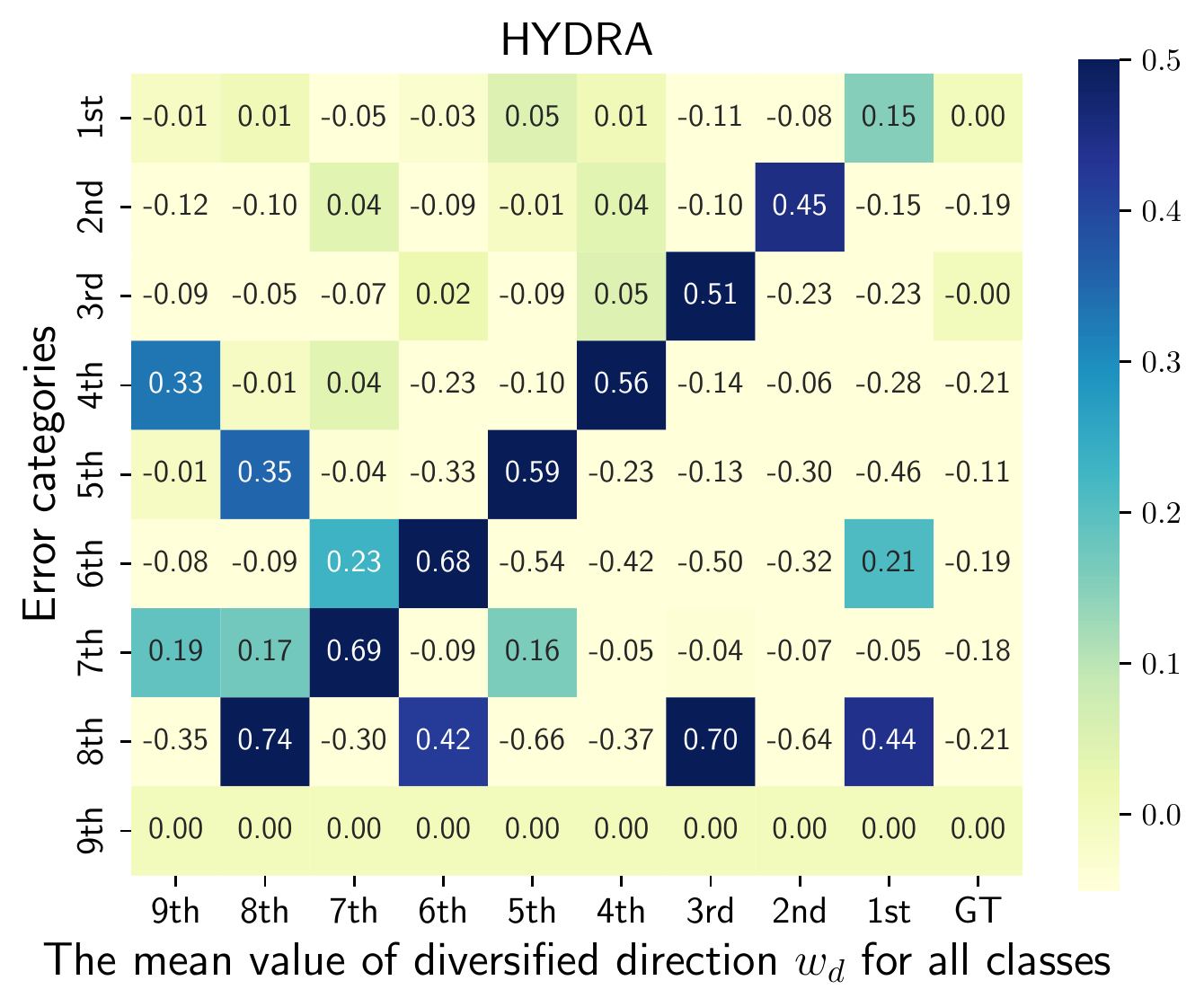}
\end{minipage}
\caption{ Quantitative statistical results of the diversified direction $\vect w_{\vect d}$ of adversarial examples on multiple defense models (~\ie, AWP~\cite{AWP}, Proxy~\cite{Proxy}, Fast~\cite{fbtf}, Feature Scatter~\cite{Feature-scatter}, Geometry~\cite{Geometry} and HYDRA~\cite{HYDRA}.). The diversified direction of each model has a model-specific bias in the positive/negative direction. In other words, random sampling is suboptimal.}
\label{fig:OSD1}
\end{figure*}

\begin{figure*}[htb]
\centering
\begin{minipage}[t]{0.45\linewidth}
\includegraphics[width=7cm]{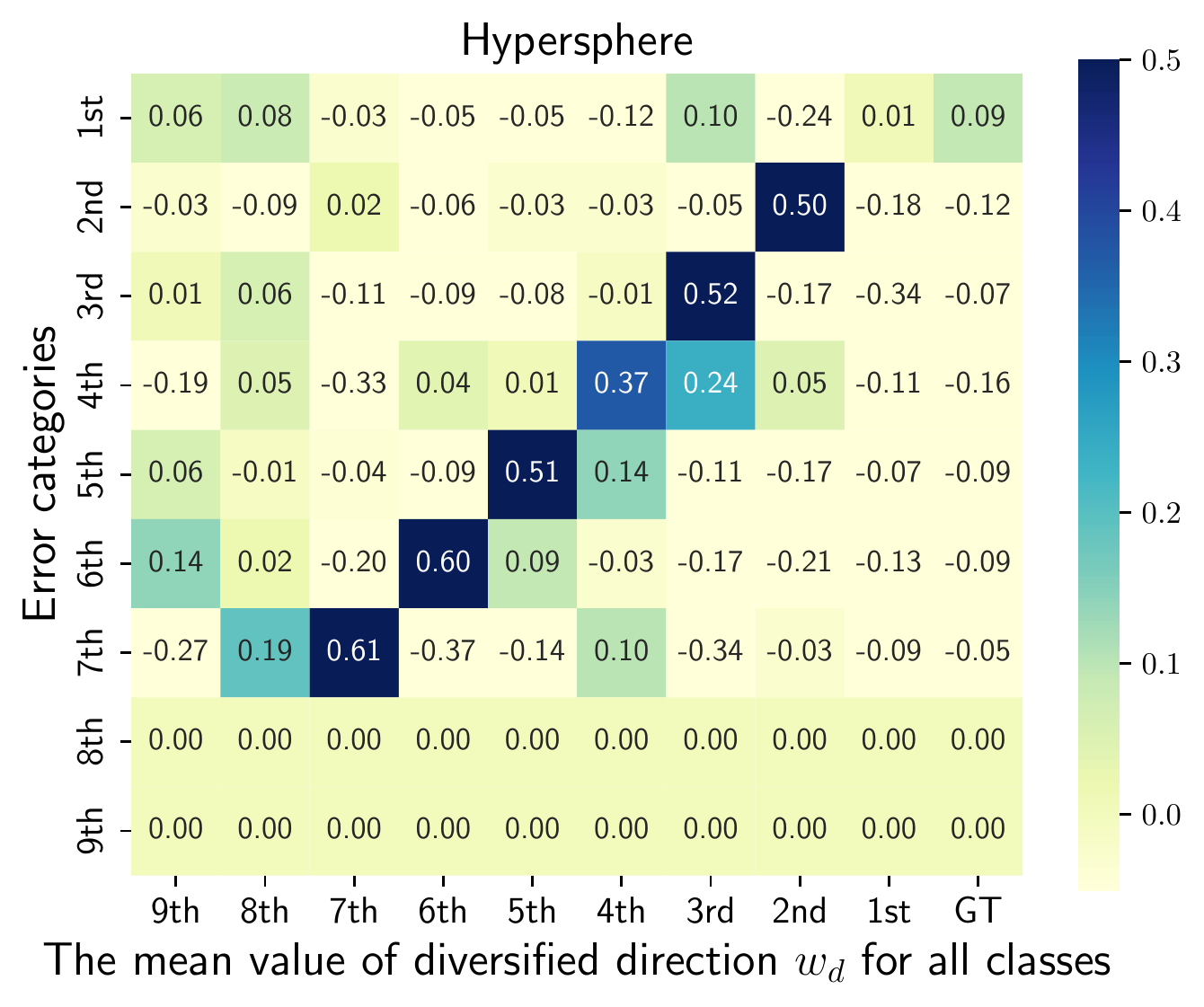}
\end{minipage}
\begin{minipage}[t]{0.45\linewidth}
\includegraphics[width=7.0cm]{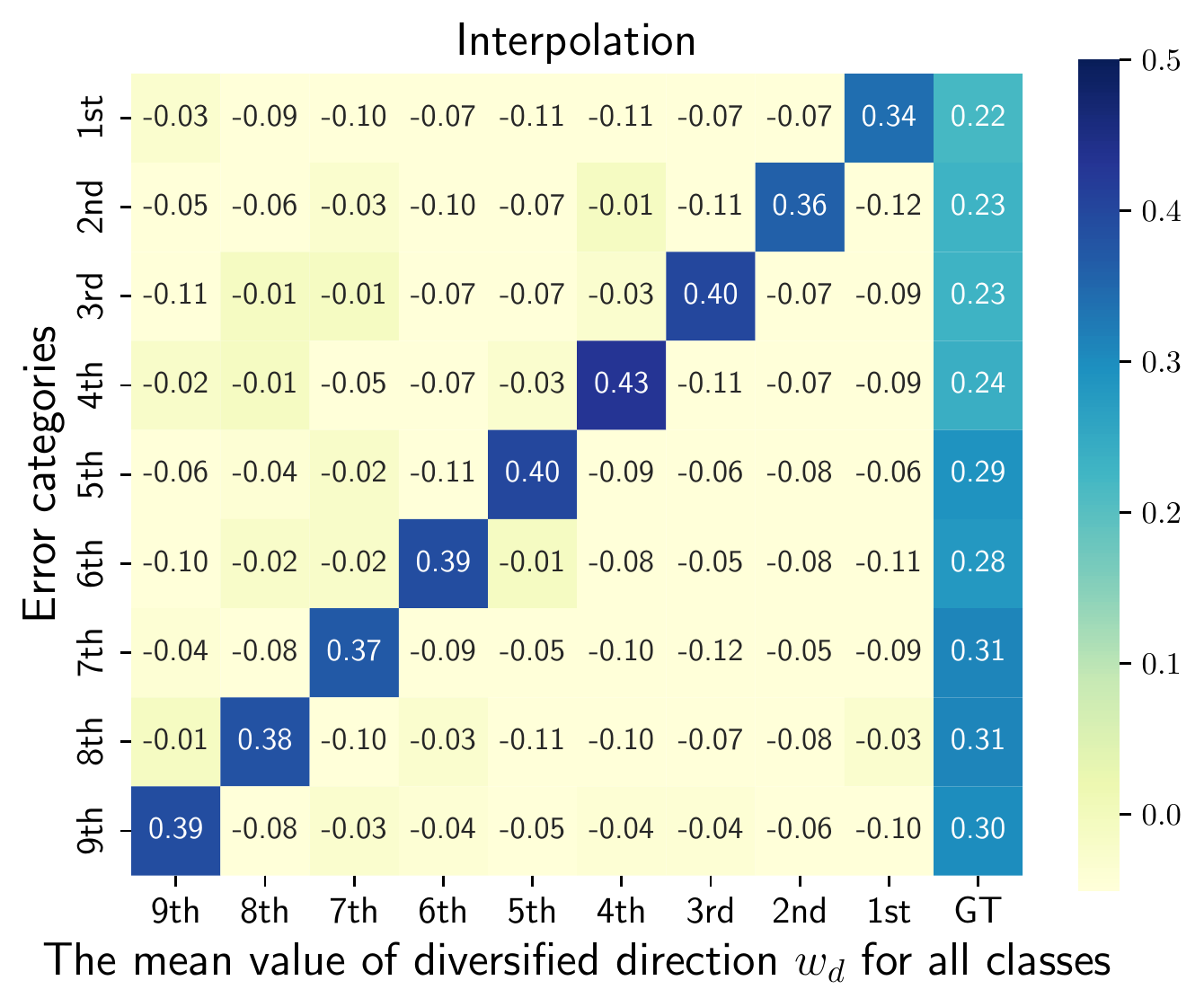}
\end{minipage}
\begin{minipage}[t]{0.45\linewidth}
\includegraphics[width=7.0cm]{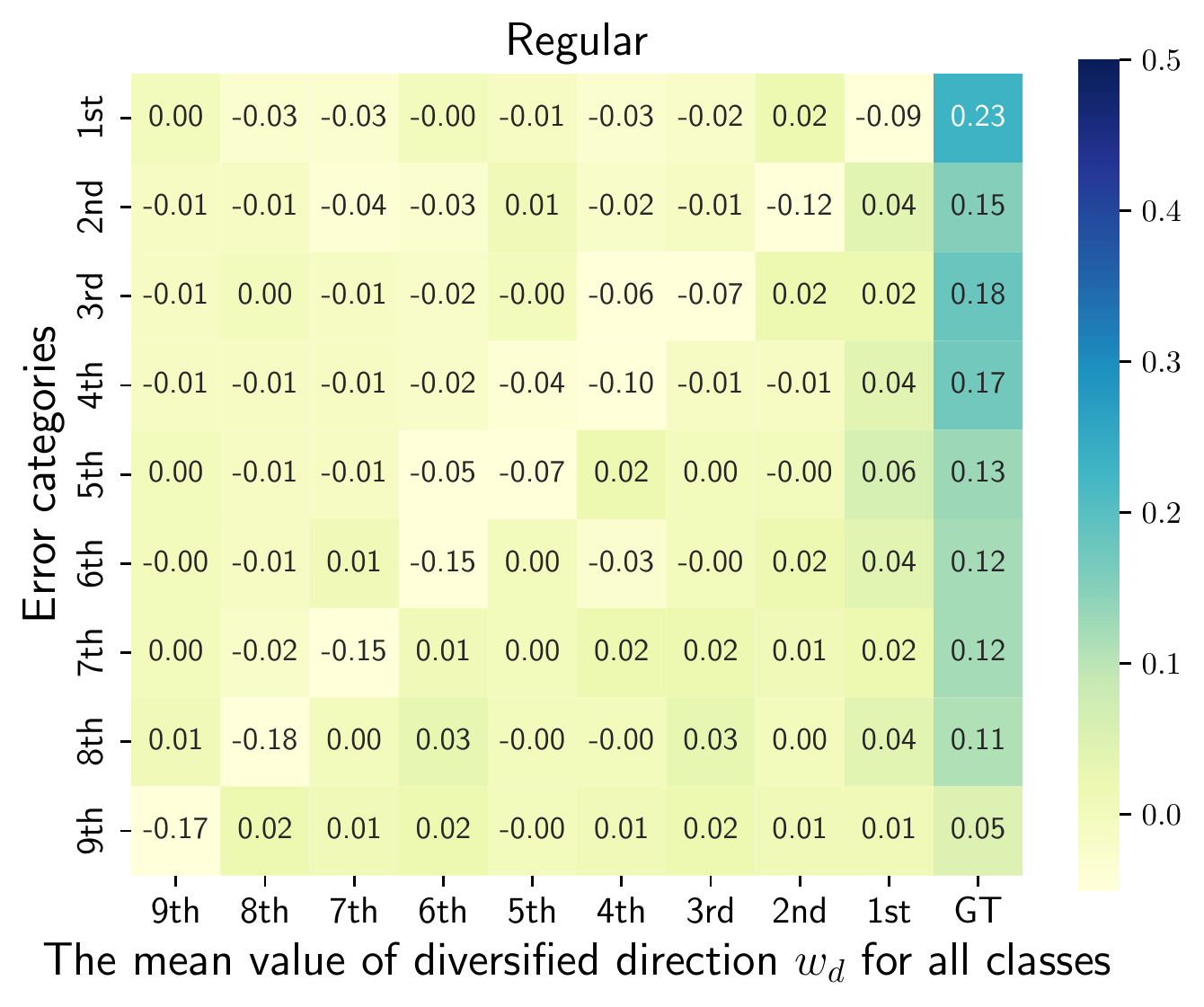}
\end{minipage}
\begin{minipage}[t]{0.45\linewidth}
\includegraphics[width=7.0cm]{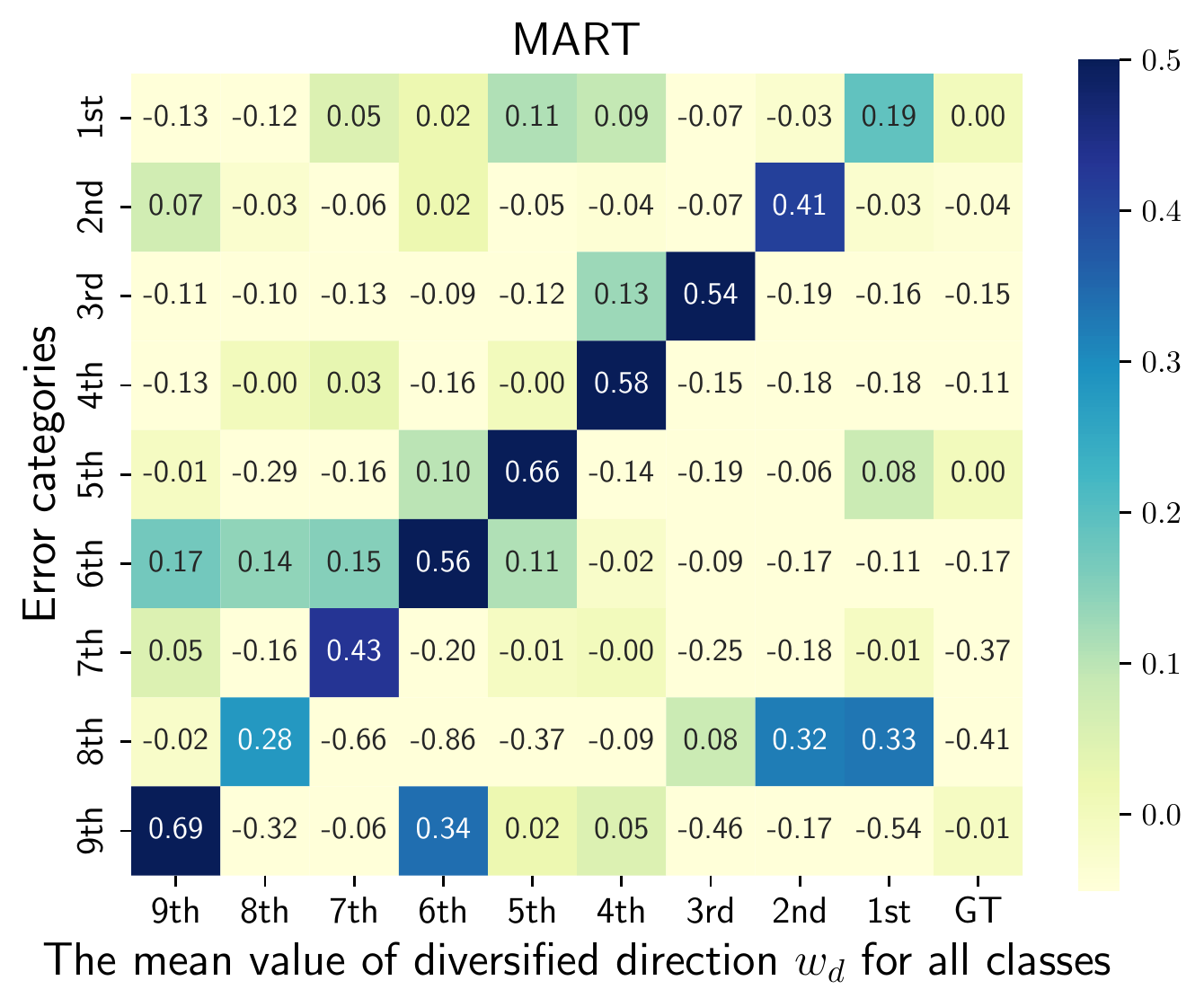}
\end{minipage}
\begin{minipage}[t]{0.45\linewidth}
\includegraphics[width=7.0cm]{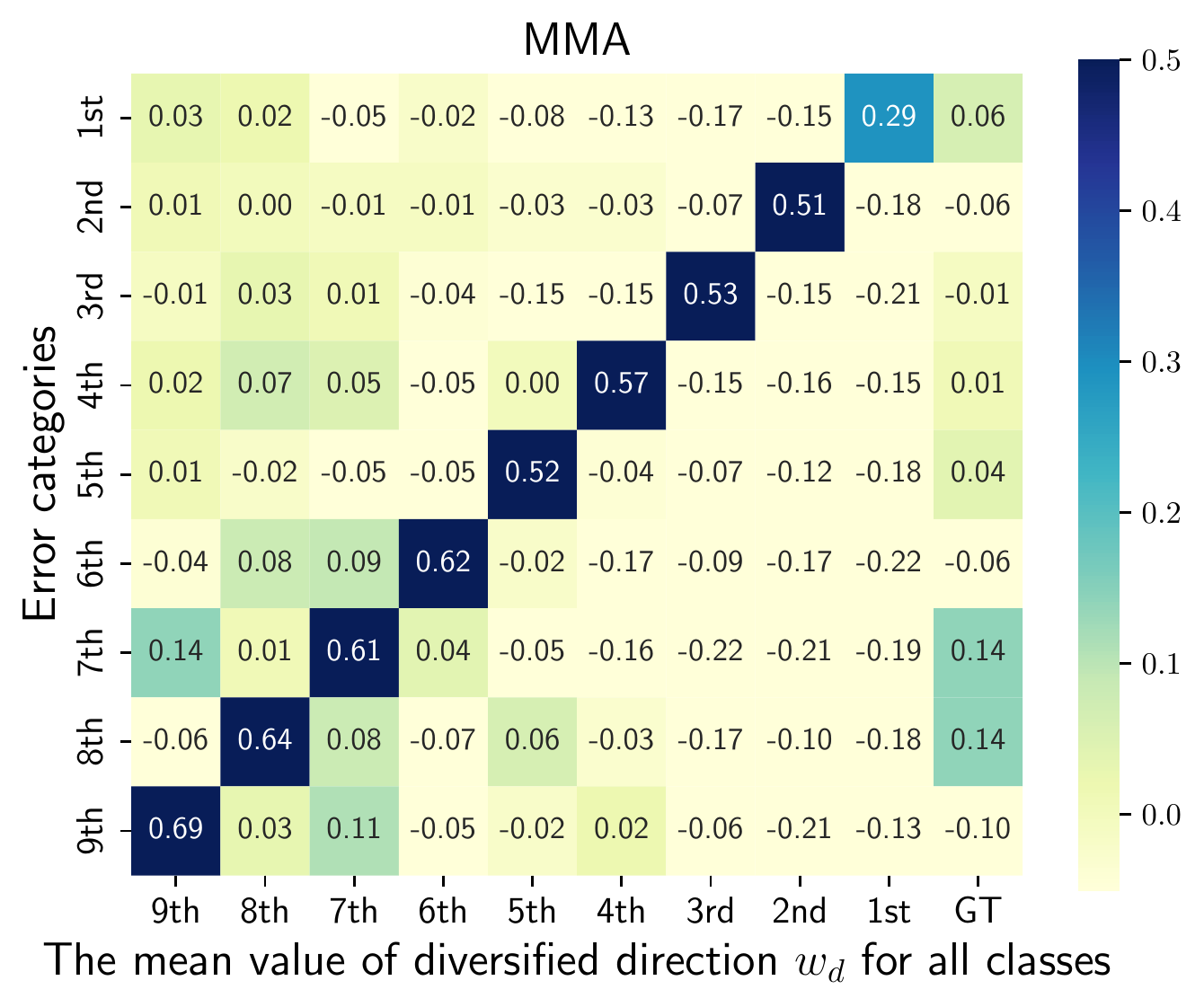}
\end{minipage}
\begin{minipage}[t]{0.45\linewidth}
\includegraphics[width=7cm]{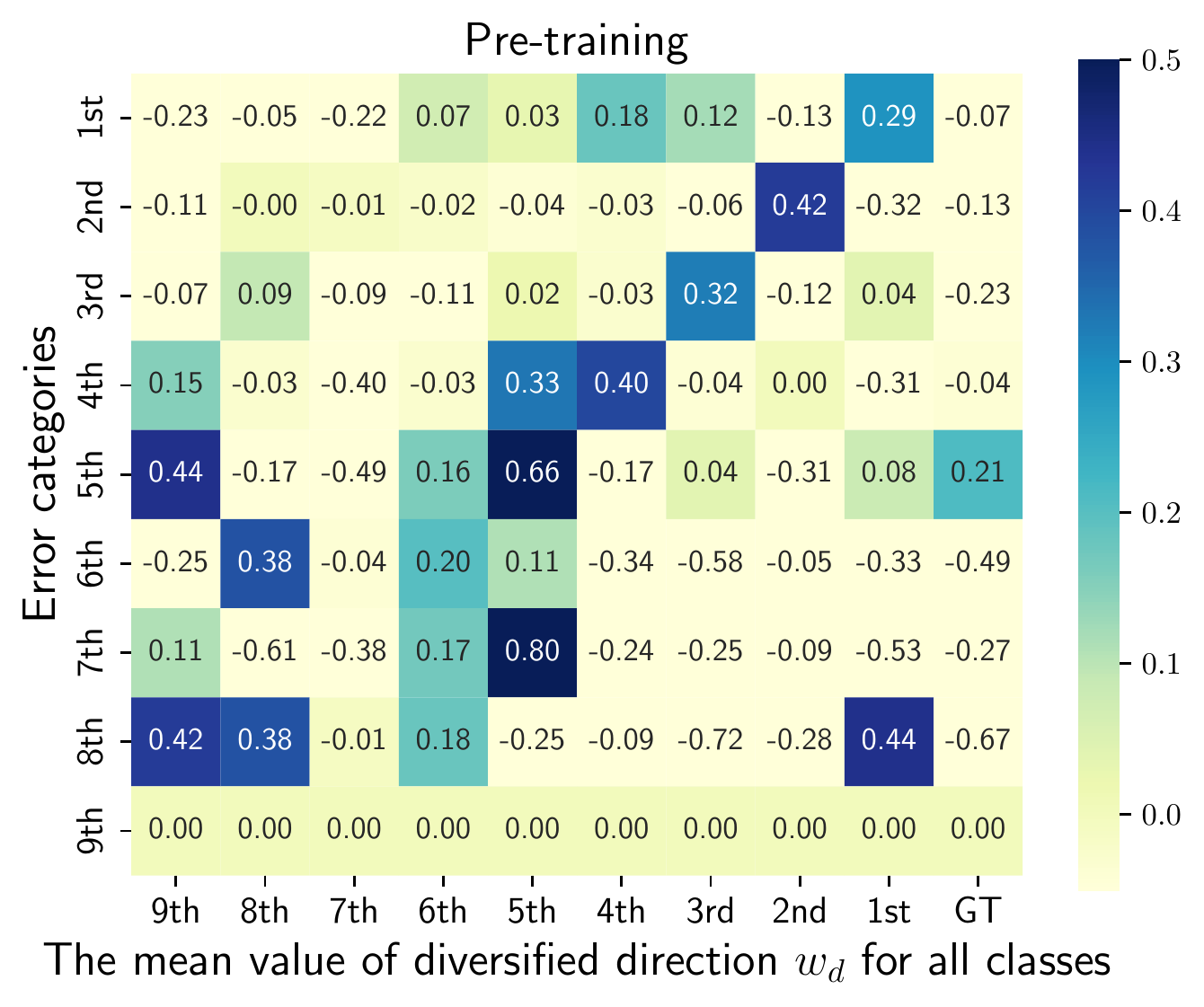}
\end{minipage}
\caption{Quantitative statistical results of the diversified direction $\vect w_{\vect d}$ of adversarial examples on multiple defense models (~\ie, Hypersphere~\cite{Hypersphere}, Interpolation~\cite{Interpolation}, Regular~\cite{Regualr}, MART~\cite{MARTs}, MMA~\cite{MMA}, Pre-training~\cite{Pre-training}.). The diversified direction of each model has a model-specific bias in the positive/negative direction. In other words, random sampling is suboptimal.}
\label{fig:OSD2}
\end{figure*}

Among adversarial examples against different models, we summarize detailed statistic results of the direction of diversification $\vect w_{\vect d}$ in \cref{fig:OSD1} and \cref{fig:OSD2}. 
For each model, there are 9 rows, representing 9 error categories, where ``1st" is the error category with the largest output logits, ``9th" is the error category with the ninth largest output logits, and so on.
There are 10 columns, representing 10 classes (9 error categories and 1 ground truth.), from ``1st" to ``9th" representing the 9 error categories and ``GT" representing the ground truth. For the error categories, we arrange the error categories in descending order according to the output logits of each error category, where the output logits refer to the output logits of the clean example corresponding to the adversarial example we counted.
The ``i" row and ``j" column represent the mean values of $\vect w_{\vect d}$ on the ``j" class when the adversarial example is misclassified as the error category with the ``i" largest output logits.
For all rows, we initialize their values to 0. We add up the $\vect{w_d}$ of all adversarial examples that are misclassified as the same row and average them. If none of the adversarial examples are misclassified as a error category, then the values of the corresponding row are 0.

From \cref{fig:OSD1} and \cref{fig:OSD2}, we have the same observations as section 3.2 of the main paper: 
1). The diversified direction $\vect {w_d}$ disobeys uniform distribution in all cases. 
2). The diversified direction for each model has a model-specific bias in the positive/negative direction, specifically, as follows:

\noindent \textbf{(a). The output logits of the error category increases, while the output logits of the ground truth decreases.} 
For most models (e.g., AWP~\cite{AWP}, Proxy~\cite{Proxy}, Fast~\cite{fbtf}, Geometry~\cite{Geometry}, HYDRA~\cite{HYDRA}, Hypersphere~\cite{Hypersphere}, MART~\cite{MARTs}, Pre-training~\cite{Pre-training}), when an adversarial example is misclassified as an error category, the $\vect {w_d}$ for the error category is mostly positive,~\ie, the output logits of the error category increases, while the $\vect {w_d}$ for the ground truth is mostly negative,~\ie, the output logits of the ground truth decreases. This is intuitive because when the output logits of the error category of adversarial examples are greater than the output logits of the ground truth, then the examples are successfully attacked.

\noindent \textbf{(b). The output logits of the error category increases, and the output logits of the ground truth also increases.} 
However, there are some models whose $\vect {w_d}$ is counter-intuitive, such as Feature Scatter~\cite{Feature-scatter},  Interpolation~\cite{Interpolation} and MMA~\cite{MMA}. When adversarial examples are misclassified as an error category, the $\vect {w_d}$ for the error category is positive,~\ie, the output logits of the error category increases, and the $\vect {w_d}$ for the ground truth is also positive,~\ie, the output logits of the ground truth also increases. Although this model has good adversarial robustness against weaker adversarial attack (~\ie, PGD~\cite{pgd}), it is poor in adversarial robustness against stronger attacks (~\ie, AA and A$^3$). A potential reason is that these defense models use gradient masks~\cite{mask}, and PGD chooses a bad starting point, which hinders the performance.

\noindent \textbf{(c). The output logits of the error category decreases, and the output logits of the ground truth also increases.} 
The most counter-intuitive is Regular~\cite{Regualr}, when an adversarial example is misclassified as an error category, the $\vect {w_d}$ for the error category is negative,~\ie, the output logits of the error category decreases, and the $\vect {w_d}$ for the ground truth is positive, ~\ie, the output logits of the ground truth increases.
This model also uses gradient masks, which leads to extremely poor adversarial robustness of this model against stronger attacks.

Since the diversified initialization directions of models have some bias, and are not uniformly distributed, generating model-specific initial directions is very important and helps to obtain better performance.

\section{Results of A$^{3}$ across various datasets, network architectures and metrics.}
In this section, we show the results of the proposed A$^{3}$ attack across various defense strategies, datasets, network architectures and metrics. The setup is the same as section 4.1 of the main paper.
\begin{table*}[h]
\vspace{-0.7em}
  \centering
  \resizebox{\textwidth}{!}{
  \begin{tabular}{@{}c|clcccccccc@{}}\toprule
  \multirow{2}{*}{\textbf{\makecell[c]{Defense Method}}}&\textbf{Dataset}&\multirow{2}{*}{\textbf{Metrics}}&\multirow{2}{*}{\textbf{Model}}&\textbf{Clean}&\multicolumn{3}{c}{\textbf{AA}}&\multicolumn{3}{c}{\textbf{A$^{3}$}}\\\cmidrule(lr){2-2}\cmidrule(lr){5-5}\cmidrule(lr){6-8}\cmidrule(lr){9-11}
                                  &  number of test samples  &   &  &acc  & acc &$\rightarrow$&$\leftarrow$& \textbf{acc} &$\rightarrow$&$\leftarrow$  \\\midrule
  \textbf{Undefended}      &ImageNet$(5000)$&$L_{\infty}(\epsilon=4/255)$         &ResNet50&$76.74$&$0.0$&$0.40$&$0.39$&$\bf{0.0}$&$\bf{0.02(20.0\times)}$&$\bf{0.005(78.0\times)}$\\
  DARI~\cite{SalmanIEKM20}      &ImageNet$(5000)$&$L_{\infty}(\epsilon=4/255)$  &WideResNet-50-2&$68.46$&$38.14$&$15.15$&$3.82$&$\bf{38.12\downarrow0.02}$&$\bf{2.67(5.67\times)}$&$\bf{1.31(2.90\times)}$\\
  DARI~\cite{SalmanIEKM20}       &ImageNet$(5000)$&$L_{\infty}(\epsilon=4/255)$         &ResNet50&$64.10$&$34.66$&$13.78$&$3.49$&$\bf{34.64\downarrow0.02}$&$\bf{2.47(5.58\times)}$&$\bf{1.22(2.86\times)}$\\
  DARI~\cite{SalmanIEKM20}       &ImageNet$(5000)$&$L_{\infty}(\epsilon=4/255)$         &ResNet18&$52.90$&$25.30$&$10.10$&$2.58$&$\bf{25.16\downarrow0.14}$&$\bf{1.96(5.15\times)}$&$\bf{0.96(2.69\times)}$\\
  DARI~\cite{SalmanIEKM20}       &ImageNet$(5000)$&$L_2(\epsilon=3.0)$               &DenseNet161&$66.14$&$36.52$&$14.51$&$3.67$&$\bf{36.50\downarrow0.02}$&$\bf{2.59(5.60\times)}$&$\bf{1.28(2.87\times)}$\\
  DARI~\cite{SalmanIEKM20}       &ImageNet$(5000)$&$L_2(\epsilon=3.0)$                  &VGG16-BN&$56.24$&$29.62$&$11.79$&$2.99$&$\bf{29.62\downarrow0.00}$&$\bf{2.20(5.36\times)}$&$\bf{1.08(2.77\times)}$\\
  DARI~\cite{SalmanIEKM20}       &ImageNet$(5000)$&$L_2(\epsilon=3.0)$                &ShuffleNet&$43.16$&$17.64$&$7.08$&$1.85$&$\bf{17.56\downarrow0.08}$&$\bf{1.58(4.48\times)}$&$\bf{0.78(2.37\times)}$\\
  DARI~\cite{SalmanIEKM20}       &ImageNet$(5000)$&$L_2(\epsilon=3.0)$              &MobileNet-V2&$49.62$&$24.78$&$9.89$&$2.52$&$\bf{24.74\downarrow0.04}$&$\bf{1.94(5.10\times)}$&$\bf{0.95(2.65\times)}$\\
  Fixing Data~\cite{fix_data}  &CIFAR10$(10000)$&$L_2(\epsilon=0.5)$         &WideResNet-28-10&$91.79$&$78.80$&$62.00$&$15.20$&$\bf{78.79\downarrow0.01}$&$\bf{5.35(11.59\times)}$&$\bf{2.63(5.78\times)}$\\
  Robustness~\cite{robustness}   &CIFAR10$(10000)$&$L_2(\epsilon=0.5)$               &ResNet50&$90.83$&$69.23$&$54.56$&$13.45$&$\bf{69.21\downarrow0.02}$&$\bf{4.72(11.56\times)}$&$\bf{2.32(5.80\times)}$\\
  Proxy~\cite{Proxy}       &CIFAR10$(10000)$&$L_2(\epsilon=0.5)$        &WideResNet-34-10&$90.31$&$76.11$&$59.89$&$14.69$&$\bf{76.10\downarrow0.01}$&$\bf{5.18(11.56\times)}$&$\bf{2.55(5.76\times)}$\\
  Overfitting~\cite{Overfitting}     &CIFAR10$(10000)$&$L_2(\epsilon=0.5)$            &ResNet18&$88.67$&$67.68$&$53.34$&$13.15$&$\bf{67.64\downarrow0.04}$&$\bf{4.61(11.57\times)}$&$\bf{2.27(5.79\times)}$\\
  ULAT~\cite{ULAT}    &MNIST$(10000)$  &$L_{\infty}(\epsilon=0.3)$     &WideResNet-28-10&$99.26$&$96.34$&$76.05$&$18.44$&$\bf{96.31\downarrow0.03}$&$\bf{6.53(11.64\times)}$&$\bf{3.22(5.71\times)}$\\
  TRADES~\cite{TRADESs}    &MNIST$(10000)$  &$L_{\infty}(\epsilon=0.3)$           &SmallCNN&$99.48$&$92.76$&$73.12$&$17.88$&$\bf{92.71\downarrow0.05}$&$\bf{6.33(11.55\times)}$&$\bf{3.12(5.73\times)}$\\\midrule
  \end{tabular}}
  \vspace{-0.7em}
  \caption{The results of the proposed A$^{3}$ attack across various defense strategies, datasets, network architectures and metrics.
    The ``acc" column shows the robust accuracies of different models.
    The ``$\rightarrow$" column shows the iteration number of forward propagation (million), while the ``$\leftarrow$" column shows the iteration number of backward propagation (million). 
    The ``\textbf{acc}" column of A$^3$ shows the difference between the robust accuracies of AA and A$^3$, the ``$\leftarrow$" and ``$\rightarrow$" columns of A$^{3}$ show the speedup factors of A$^3$ relative to AA.}
    \label{table.main}
  \end{table*}

\noindent{\textbf{Results.}} As can be seen in \cref{table.main}. we show the effectiveness of the proposed A$^3$ across more datasets (e.g., MNIST, CIFAR10, and ImageNet), network architectures (e.g., VGG16, DenseNet161, ShuffleNet, etc.) and metrics (e.g., $L_{\infty}$ and $L_2$).
The experimental results show that A$^3$ is better than AA on various datasets, model architectures and metrics.

\end{document}